\pdfoutput=1
\documentclass[11pt]{article}

\usepackage{graphicx} % Required for inserting images
\usepackage{todonotes}
\usepackage{xargs}    
\usepackage{tikz}
\usepackage{fontawesome5}
\usetikzlibrary{shadows,arrows,decorations.pathreplacing,positioning,shapes,calc,fit}
\usetikzlibrary{shapes,arrows.meta,positioning}
\usepackage{amsmath}
\usepackage{wasysym}
\usepackage{amssymb}
\usepackage{subcaption}
\usepackage{svg}
\usepackage{booktabs}
\usepackage{multirow}
\usetikzlibrary{decorations.pathreplacing,calc}

\usepackage{xcolor} % Needed for the postbreak marker colors
\usepackage{listings}
\usepackage{tcolorbox}
\tcbuselibrary{breakable,skins,theorems}   % ← no spaces
\definecolor{strategycolor_1pass}{HTML}{55A868} % Corresponds to the hex '#55A868'
\definecolor{strategycolor_1by1}{HTML}{C44E52} % Corresponds to the hex '#C44E52'
\usepackage{tikz}
\usetikzlibrary{arrows,decorations.pathreplacing,positioning,shapes,calc,fit}
\usepackage[preprint]{acl}

\usepackage{times}
\usepackage{latexsym}

\usepackage[T1]{fontenc}
\usepackage[utf8]{inputenc}

\usepackage{microtype}

\usepackage{inconsolata}

\usepackage{graphicx}

\usepackage{color,soul}
\usepackage{enumitem}

\title{
Can AI Validate Science \tikz[baseline]{\node[fill=purple!10,rounded corners=2pt,anchor=base] {?};} \raisebox{-1.4ex}{\includegraphics[height=2.5em]{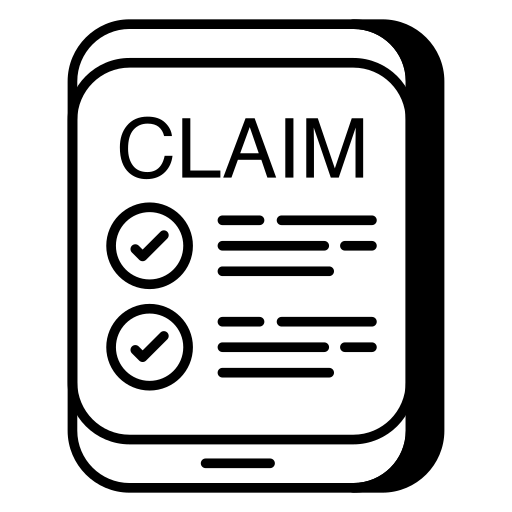}}\\
Benchmarking LLMs for Accurate Scientific \\
\tikz[baseline]{\node[fill=blue!10,rounded corners=2pt,anchor=base] (c) {Claim};}
$\boldsymbol{\rightarrow}$
\tikz[baseline]{\node[fill=green!10,rounded corners=2pt,anchor=base] (e) {Evidence};}
Reasoning
}

\author{Shashidhar Reddy Javaji, Yupeng Cao, Haohang Li, Yangyang Yu \\
{\bf Nikhil Muralidhar}, {\bf Zining Zhu}  \\
Stevens Institute of Technology}
\begin{document}
\maketitle
\begin{abstract}
Large language models (LLMs) are increasingly being used for complex research tasks such as literature review, idea generation, and scientific paper analysis, yet their ability to truly understand and process the intricate relationships within complex research papers, such as the logical links between claims and supporting evidence remains largely unexplored.
In this study, we present CLAIM-BENCH, a comprehensive benchmark for evaluating LLMs' capabilities in scientific claim-evidence extraction and validation, a task that reflects deeper comprehension of scientific argumentation. We systematically compare three approaches 
 which are inspired by divide and conquer approaches, across six diverse LLMs, highlighting model-specific strengths and weaknesses in scientific comprehension. Through evaluation involving over 300 claim-evidence pairs across multiple research domains, we reveal significant limitations in LLMs' ability to process complex scientific content. Our results demonstrate that closed-source models like GPT-4 and Claude consistently outperform open-source counterparts in precision and recall across claim-evidence identification tasks. Furthermore, strategically designed three-pass and one-by-one prompting approaches significantly improve LLMs' abilities to accurately link dispersed evidence with claims, although this comes at increased computational cost. CLAIM-BENCH sets a new standard for evaluating scientific comprehension in LLMs, offering both a diagnostic tool and a path forward for building systems capable of deeper, more reliable reasoning across full-length papers. \footnotemark
 \footnotetext{To facilitate future research and standardize evaluation in this area, we release \textsc{CLAIM-BENCH} at \href{https://github.com/shashidharjavaji/RC_BENCH}{\textit{the CLAIM\_BENCH GitHub repository}}.}

 % , providing a comprehensive framework for assessing and improving LLMs' scientific paper processing capabilities.

% \improvement{Maybe the name of the Benchmark needs some revision}

\end{abstract}

\section{Introduction}
% \todo{ZZ: Can we perhaps reorganize the Introduction following this workflow: (1) AI4Research is good but people don't know how good they are. (2) For AIs to do AI4Research, they need to perform many tasks. We are interested in one gap that is under-explored: the claim-evidence reasoning. (3) Claim-evidence reasoning requires high-level understanding and reasoning. We need to benchmark these capabilities. (4) Therefore, we present CLAIM-Bench.}

Large Language Models (LLMs) have become important tool in academic research, demonstrating impressive capabilities such as automating comprehensive literature reviews, facilitating innovative idea generation, and aiding experimental design. These advancements promise significant improvements in research productivity, creativity, and efficiency, fueling excitement about the transformative potential of AI-driven methodologies in science. However, as researchers increasingly assign critical tasks to these models—from content summarization and hypothesis generation to automated peer review \citep{checco_ai-assisted_2021, agarwal_litllm:_2025, lu_ai_2024}—a fundamental yet overlooked question emerges: how deeply do these models truly understand scientific knowledge beyond surface-level pattern recognition? Despite their widespread use and promising outcomes, there remains uncertainty about the depth and accuracy of their reasoning capabilities, particularly in complex scientific contexts.

Scientific papers are characterized by intricate relationships, primarily structured around claims supported by corresponding evidence. The ability to accurately identify and reason about these claim-evidence pairs is essential for validating scientific findings and ensuring research integrity, making it a critical test of LLMs’ comprehension depth. Unlike surface-level tasks such as summarization or question answering, claim-evidence identification requires global reasoning across paper sections, synthesis of dispersed information, and a nuanced understanding of logical dependencies. While existing works have assessed LLMs' capabilities in related research tasks such as summarization \citep{agarwal_litllm:_2025}, literature synthesis \citep{lu_ai_2024}, and hypothesis generation \citep{vladika_scientific_2023}, none have explicitly benchmarked LLM performance on systematically extracting and validating claims with supporting evidence, leaving this area of scientific comprehension underexplored.

Despite the importance of accurately reasoning about claims and supporting evidence, no existing benchmarks explicitly assess LLM capabilities for this specific type of high-level scientific reasoning. Benchmarks such as LongGenBench \citep{wu_longgenbench:_2025} and XL2Bench \citep{ni_xl$^2$bench:_2024} have highlighted persistent limitations in LLMs' abilities to process long-context inputs and maintain logical coherence. Similarly, peer review frameworks like MetaWriter \citep{sun_metawriter:_2024} and AGENTREVIEW \citep{jin_agentreview:_2024} evaluate LLMs in automated review contexts but do not specifically test their capability to validate logical relationships such as claims and evidence, a task crucial for rigorous scientific evaluation. Findings from Chain of Evidence (CoE) frameworks \citep{chang_what_2024} underscore the complexity of structured, multi-hop reasoning required to integrate and validate information dispersed across documents. All these works evaluate reasoning in the general domains, but the scientific reasoning capability, which imposes unique challenges, is not benchmarked.

% \ZZ{There are some scientific benchmarks but these benchmarks are not specific enough for C-E (the current paragraph argues backwards. C-E is more specific than the ScienceAgentBench)} 
% Specific to the scientific reasoning, recent works including The AI Scientist \citep{lu_ai_2024}, LitLLM \citep{agarwal_litllm:_2025}, and ChatCite \citep{li_chatcite:_2025}, have indeed benchmarked various reasoning tasks relevant to scientific workflows. Likewise, frameworks such as ScienceAgentBench \citep{chen_scienceagentbench:_2025} and SCBENCH \citep{li_scbench:_2025} focus on multi-step reasoning and long-context understanding. However, these evaluations remain primarily localized or confined to predefined tasks, leaving critical gaps in systematically assessing the nuanced, cross-sectional reasoning required for validating claims with dispersed evidence throughout full-length scientific papers.

Within scientific reasoning, The AI Scientist \citep{lu_ai_2024}, LitLLM \citep{agarwal_litllm:_2025}, and ChatCite \citep{li_chatcite:_2025} benchmark LLMs on tasks such as literature review and hypothesis generation, while ScienceAgentBench \citep{chen_scienceagentbench:_2025} and SCBENCH \citep{li_scbench:_2025} probe multi-step reasoning and long-context understanding. However, none of these frameworks explicitly measure the finer-grained ability to verify whether the evidence presented in a full scientific paper truly supports its claims—precisely the claim-and-evidence (C-E) reasoning capability our benchmark targets.

To address these gaps, we present CLAIM-BENCH, a novel benchmark designed to systematically evaluate LLMs' abilities to identify and validate claim-evidence relationships in scientific papers. CLAIM-BENCH challenges LLMs to process entire scientific papers, connect ideas across sections, and reason about them on a high level. In this work, we evaluate six state-of-the-art LLMs across diverse research domains. Our experiments indicate that larger models (e.g., GPT-4-Turbo, Claude 3.5) maintain high recall even with lengthy documents, especially when using iterative prompting, whereas smaller models (e.g., LLaMA, Ministral) experience significant performance drops with increasing document length specially under Single-Pass prompting. These findings highlight crucial areas for enhancing long-context comprehension and inform the development of reliable AI-driven tools for scientific research and peer review.

% To clearly address these gaps and systematically evaluate LLM capabilities, we propose the following key research questions

% \begin{itemize}[nosep]
% \item To what extent can LLMs accurately identify claims and their corresponding evidence across entire research papers?
% \item How do different prompting strategies—Single-Pass, multi-pass iterative, and three-pass strategic approaches—affect the performance of LLMs on claim-evidence identification tasks? (We discuss the strategies in Section~\ref{analysis_methods})
% \item What are the limitations in existing LLM architectures that prevent effective reasoning and logical synthesis across long scientific texts?
% \end{itemize}

\section{Related Work}

% \ZZ{Group the paragraphs in this section with the $\backslash$paragraph. The organized paragraphs should appear similar to Shravan's paper}
\paragraph{AI for Science} Large Language Models (LLMs) have significantly advanced scientific workflows, facilitating tasks such as peer review and hypothesis generation. Tools like ReviewerGPT \citep{liu_reviewergpt?_2023} and ReviewFlow \citep{sun_reviewflow:_2024} have streamlined peer review processes, while AGENTREVIEW \citep{jin_agentreview:_2024} simulates collaborative review systems to improve research evaluation workflows. In parallel, fact-checking frameworks, such as Scientific Fact-Checking \citep{vladika_scientific_2023} and Exploring Multidimensional Checkworthiness \citep{liu_exploring_2025}, emphasize validating claims in scientific literature. However, these systems primarily focus on localized tasks or prioritization mechanisms, leaving the broader challenge of understanding the connections across entire documents by LLMs unaddressed. Additional work such as AI-assisted peer review \citep{checco_ai-assisted_2021} explores the feasibility of algorithmically approximating peer-review judgments, raising key ethical and practical concerns.

\paragraph{Benchmarks}
Long-context benchmarks, such as SCBENCH \citep{li_scbench:_2025}, MMLongBench-Doc \citep{ma_mmlongbench-doc:_2024}, and LongGenBench \citep{wu_longgenbench:_2025}, have assessed LLMs' ability to process extended inputs and maintain coherence, focusing primarily on tasks like document summarization and long-form generation. Specialized benchmarks like U-MATH \citep{chernyshev_u-math:_2025} and Leave No Document Behind \citep{godbole_leveraging_2024} examine domain-specific reasoning and multi-document synthesis but address relatively structured and localized relationships.  The LCFO benchmark \citep{costa-jussa_lcfo:_2024} targets summary expansion with varying granularities of content compression, revealing limits in semantic retention.
The Y-NQ dataset \citep{costa-jussa_y-nq:_2024} exposes disparities in open-book comprehension across low- \& high-resource languages, hinting at deeper weaknesses in cross-lingual and low-resource long-context understanding.
Data Interpreter \citep{hong_data_2024} showcases long-term data analysis workflows with LLM agents, but primarily focuses on task planning and execution rather than deep textual reasoning.
In neuroscience, \citep{luo_large_2025} show LLMs surpassing expert predictions in future experimental outcomes, yet such success doesn't imply comprehension of reasoning chains.
In contrast, our work focuses specifically on research papers, which are characterized by more complex and dispersed relationships, such as claims supported by evidence across multiple sections. CLAIM-BENCH evaluates the ability of LLMs to synthesize these intricate connections, testing their capacity for global reasoning and coherence in a way that reflects the unique demands of scientific texts.

\paragraph{Collaborative Reasoning}
Collaborative reasoning frameworks offer a complementary perspective, with multi-agent systems like Two Heads Are Better Than One \citep{su_many_2025} and iterative feedback mechanisms such as CYCLERESEARCHER \citep{weng_cycleresearcher:_2025} showing promise in enhancing reasoning capabilities. While these approaches address some limitations of Single-Pass LLM systems, their primary focus remains on generating and refining content rather than validating complex logical relationships. Similarly, tools like AIGS \citep{liu_aigs:_2024} and LLM-Assisted Hypothesis Generation \citep{vladika_scientific_2023} explore reasoning and hypothesis testing but do not directly tackle the problem of scientific comprehension. \citep{leng_llm-assisted_2024} introduce a graph-based approach for hypothesis generation and evaluation, demonstrating potential for structured creativity, yet falling short of validating interlinked arguments at scale.

\paragraph{Ethical AI} Finally, ethical considerations have been raised in works like Ethical Use of LLMs \citep{lissack_ethical_2024}, which stresses the need for transparency and accountability in AI-driven research, and multimodal benchmarks like MileBench \citep{dingjie_milebench:_2024}, which expand the scope of LLM evaluation to include visual and textual data. These efforts, while addressing important aspects of AI integration in research, highlight the absence of targeted benchmarks that evaluate claim-evidence validation across long, complex scientific texts—a gap CLAIM-BENCH aims to fill.

% \subsection{Long-Context}

% \subsection{AI for Science}

% \subsection{Existing Benchmarks in AI for Research/Science}

\section{Methodology}

% In this study, we introduce CLAIM-BENCH, a benchmark framework to evaluate LLMs effectiveness in identifying and analyzing claim-evidence relationships within academic texts. This framework is designed to address the significant challenges posed by lengthy academic texts and the identification of logical connections between claims and supporting evidence. Through this approach, we aim to enhance both the theoretical understanding of LLM capabilities and their practical applications in automated literature analysis and information extraction.
In this section, we present the design of CLAIM-BENCH, our benchmark for evaluating how well LLMs identify and analyze claim–evidence relationships in full-length research papers. 

% \ZZ{If we run out of page limits, remove the rest of this paragraph (metadiscourse).} Our methodology is systematically structured into key subsections. It begins with dataset creation (Section \ref{Dataset}), followed by model selection (Section \ref{Model_selection}), where we outline the criteria for choosing specific LLMs. Subsequent sections delve into our analysis methods (Section \ref{analysis_methods}), which focus on various prompting techniques to gauge their impact on model performance. Finally, we discuss the evaluation metrics (Section \ref{Evaluation_metrics}) used in this study, elaborating on why certain metrics were chosen to assess the effectiveness of LLMs in our benchmark tests. This structured approach illuminates current LLM capabilities while establishing a foundation for future research advancements.

\subsection{Dataset}
\label{Dataset}

\paragraph{Dataset Curation}
The dataset for this study was curated by 4 PhD students with research experience. Each annotator had at least one first-author conference publication, ensuring familiarity with scientific writing standards. These researchers selected papers according to specific guidelines (Appendix \ref{Annotator_guidelines}) to ensure relevance and diversity. Selection criteria included: papers from the year 2024, non-math-intensive subjects, length between 0 to 20 pages. The aim was to represent a broad spectrum of current AI/ML research topics within the dataset. 

% \todo{Need to make changes in this section} 

To facilitate easier annotations, we developed a PDF annotation tool, it lets users load a paper, drag a pointer over any sentence or paragraph to mark it as a claim, then click-add evidence additional spans as linked evidence for that claim; each claim–evidence pair is stored in a one-to-many structure and exported as JSON. (see Appendix \ref{annotation_tool}). 

\paragraph{Annotation Quality Check}
After compiling the initial annotations (100 papers), these were set aside before evaluating the models to ensure an unbiased assessment of their capabilities. To enhance the reliability of our dataset as ground truth, we conducted a validation phase where a different set of annotators re-annotated a subset of 30 papers.\noindent
We measured inter–annotator agreement with two metrics.  
\textbf{(1) F1:} Averaging symmetric $F_1$ across annotator pairs gives substantial agreement for claims ($0.755$) and moderate agreement for evidence ($0.659$) and claim–evidence links ($0.617$).  
\textbf{(2) Cohen’s $\kappa$:} Averaging $\kappa$ across pairs yields $0.66$ for claims (substantial) and $0.30$ for evidence (fair).  
Together, these scores confirm that \textsc{CLAIM-BENCH} is a reliable yet challenging benchmark (details in 
Appendix~\ref{inter_annotator_agreement}).

% \paragraph{Dataset statistics}
% \ZZ{Briefly mention some dataset statistics: how many papers does RC-Bench have? How many claims are there per paper? How many evidences are there? On average, how many evidences does each claim have?}

% \ZZ{Somewhere in this subsection, mention whether the claim or the evidence is a complete sentence, (or if they are just text spans). If they are not complete sentences, it'd be great to also include the number of words per C span or per E span.}

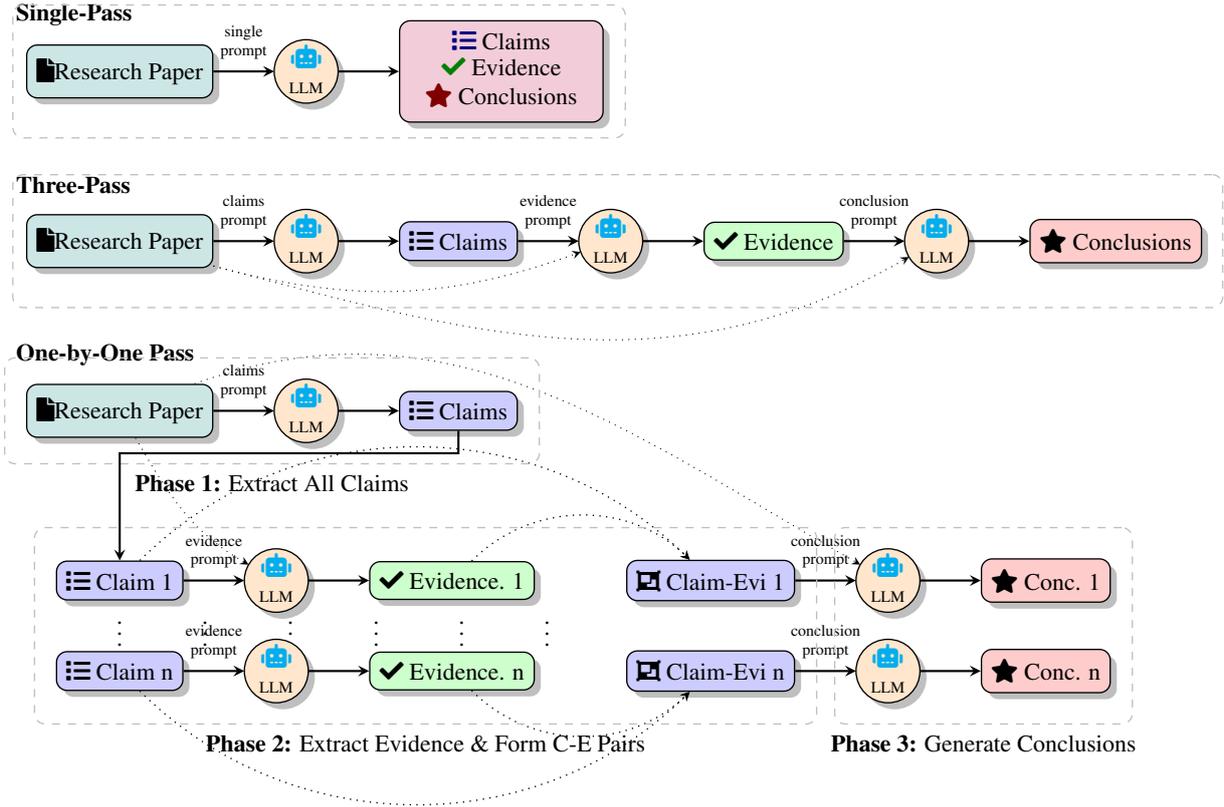
\begin{figure*}[t]
\centering
\begin{tikzpicture}[
    scale=0.75,
    >=stealth,
    line width=0.5pt,
    node distance=0.8cm,
    paper/.style={rectangle, draw, fill=teal!20, minimum width=1.6cm, minimum height=0.7cm, rounded corners, font=\footnotesize, drop shadow},
    llm/.style={circle, draw, fill=orange!20, minimum size=0.4cm, inner sep=0pt, outer sep=0pt, font=\footnotesize, drop shadow},
    claim/.style={rectangle, draw, fill=blue!20, minimum width=1cm, minimum height=0.5cm, rounded corners, font=\footnotesize, drop shadow},
    evidence/.style={rectangle, draw, fill=green!20, minimum width=1cm, minimum height=0.5cm, rounded corners, font=\footnotesize, drop shadow},
    conclusion/.style={rectangle, draw, fill=red!20, minimum width=1cm, minimum height=0.5cm, rounded corners, font=\footnotesize, drop shadow},
    results/.style={rectangle, draw, fill=purple!20, minimum width=1.1cm, minimum height=1.1cm, rounded corners, font=\footnotesize, drop shadow},
    phase/.style={draw=gray!50, dashed, rounded corners, inner sep=8pt},
    arrow/.style={->, thick, scale=0.5},
    curved/.style={->, dotted, bend left=30, scale=0.5}
]

% Case 1: Single-Pass
\begin{scope}[yshift=6cm]
\node[anchor=west, font=\footnotesize] at (-2,1) {\textbf{Single-Pass}};

\node[paper] (paper1) at (0,0) {\shortstack{\faFile Research Paper}};
\node[llm, right=0.8cm of paper1] (llm1) {\begin{tabular}{@{}c@{}}\raisebox{-0.5pt}{\textcolor{cyan}{\small\faRobot}} \\ \raisebox{-1pt}{\tiny LLM}\end{tabular}};
\node[results, right=0.8cm of llm1] (output1) {\begin{tabular}{c} 
    {\color{blue!50!black}\faList} Claims \\ 
    {\color{green!50!black}\faCheck} Evidence \\ 
    {\color{red!50!black}\faStar} Conclusions 
\end{tabular}};

\draw[arrow] (paper1) -- node[above, font=\tiny] {\begin{tabular}{c} single \\ prompt \end{tabular}} (llm1);
\draw[arrow] (llm1) -- (output1);

\node[phase, fit={($(paper1)+(-1.5,0.8)$) ($(output1)+(1.8,-0.8)$)}] (case1box) {};
%\node[above, font=\footnotesize] at (case1box.north) {\textbf{Single-Pass Approach}};
\end{scope}

% Case 2: Three-Pass
\begin{scope}[yshift=3cm]
\node[anchor=west, font=\footnotesize] at (-2,1) {\textbf{Three-Pass}};

\node[paper] (paper2) at (0,0) {\shortstack{\faFile Research Paper}};
\node[llm, right=0.8cm of paper2] (llm2a) {\begin{tabular}{@{}c@{}}\raisebox{-0.5pt}{\textcolor{cyan}{\small\faRobot}} \\ \raisebox{-1pt}{\tiny LLM}\end{tabular}};
\node[claim, right=0.8cm of llm2a] (claims2) {\faList\ Claims};
\node[llm, right=0.8cm of claims2] (llm2b) {\begin{tabular}{@{}c@{}}\raisebox{-0.5pt}{\textcolor{cyan}{\small\faRobot}} \\ \raisebox{-1pt}{\tiny LLM}\end{tabular}};
\node[evidence, right=0.8cm of llm2b] (evidence2) {\faCheck\ Evidence};
\node[llm, right=0.8cm of evidence2] (llm2c) {\begin{tabular}{@{}c@{}}\raisebox{-0.5pt}{\textcolor{cyan}{\small\faRobot}} \\ \raisebox{-1pt}{\tiny LLM}\end{tabular}};
\node[conclusion, right=0.8cm of llm2c] (conclusions2) {\faStar\ Conclusions};

% Primary arrows for the main flow
\draw[arrow] (paper2) -- node[above, font=\tiny] {\begin{tabular}{c} claims \\ prompt \end{tabular}} (llm2a);
\draw[arrow] (llm2a) -- (claims2);
\draw[arrow] (claims2) -- node[above, font=\tiny] {\begin{tabular}{c} evidence \\ prompt \end{tabular}} (llm2b);
\draw[arrow] (llm2b) -- (evidence2);
\draw[arrow] (evidence2) -- node[above, font=\tiny] {\begin{tabular}{c} conclusion \\ prompt \end{tabular}} (llm2c);
\draw[arrow] (llm2c) -- (conclusions2);

% Added dotted connections from paper to each LLM with arrows and lower curves
\draw[->, dotted, black, thin] (paper2) to[out=-15,in=200] (llm2b);
\draw[->, dotted, black, thin] (paper2) to[out=-15,in=210] (llm2c);

\node[phase, fit={($(paper2)+(-1.5,0.8)$) ($(conclusions2)+(1.5,-0.8)$)}] (case2box) {};
\end{scope}

% Case 3: One-by-One Pass
\node[anchor=west, font=\footnotesize] at (-2,1) {\textbf{One-by-One Pass}};

% Phase 1
\begin{scope}[local bounding box=phase1]
\node[paper] (paper3) at (0,0) {\shortstack{\faFile Research Paper}};
    \node[llm, right=0.8cm of paper3] (llm3) {\begin{tabular}{@{}c@{}}\raisebox{-0.5pt}{\textcolor{cyan}{\small\faRobot}} \\ \raisebox{-1pt}{\tiny LLM}\end{tabular}};
    \node[claim, right=0.8cm of llm3] (allclaims) {\faList\ Claims};
    
    \draw[arrow] (paper3) -- node[above, font=\tiny] {\begin{tabular}{c} claims \\ prompt \end{tabular}} (llm3);
    \draw[arrow] (llm3) -- (allclaims);
\end{scope}

% Phase 2 with increased gap
\begin{scope}[yshift=-3cm]
    \node[claim] (claim1) at (0,0) {\faList\ Claim 1};
    \node[llm, right=0.8cm of claim1] (llm4) {\begin{tabular}{@{}c@{}}\raisebox{-0.5pt}{\textcolor{cyan}{\small\faRobot}} \\ \raisebox{-1pt}{\tiny LLM}\end{tabular}};
    \node[evidence, right=0.8cm of llm4] (evidence1) {\faCheck\ Evidence. 1};
    \node[claim, right=1.2cm of evidence1] (pair1) {\faObjectGroup\ Claim-Evi 1};
    
    \node[font=\footnotesize] at (0,-0.8) {$\vdots$};
    \node[font=\footnotesize] at (1.5,-0.8) {$\vdots$};
    \node[font=\footnotesize] at (3,-0.8) {$\vdots$};
    \node[font=\footnotesize] at (4.5,-0.8) {$\vdots$};
    
    \node[claim] (claimn) at (0,-1.6) {\faList\ Claim n};
    \node[llm, right=0.8cm of claimn] (llm5) {\begin{tabular}{@{}c@{}}\raisebox{-0.5pt}{\textcolor{cyan}{\small\faRobot}} \\ \raisebox{-1pt}{\tiny LLM}\end{tabular}};
    \node[evidence, right=0.8cm of llm5] (evidencen) {\faCheck\ Evidence. n};
    \node[claim, right=1.2cm of evidencen] (pairn) {\faObjectGroup\ Claim-Evi n};
    
    \draw[arrow] (claim1) -- node[above, font=\tiny] {\begin{tabular}{c} evidence \\ prompt \end{tabular}} (llm4);
    \draw[arrow] (llm4) -- (evidence1);
    \draw[arrow] (claimn) -- node[above, font=\tiny] {\begin{tabular}{c} evidence \\ prompt \end{tabular}} (llm5);
    \draw[arrow] (llm5) -- (evidencen);
    
    % Added dotted connections from paper to evidence extraction LLMs with arrows and lower curves
    \draw[->, dotted, black, thin] (paper3) to[out=-60,in=150] (llm4);
    % \draw[->, dashed, gray, thin] (paper3) to[out=-70,in=150] (llm5);
    
    \draw[curved] (claim1) to[out=45, in=135] (pair1);
    \draw[curved] (evidence1) to[out=45, in=135] (pair1);
    \draw[curved] (claimn) to[out=-45, in=-135] (pairn);
    \draw[curved] (evidencen) to[out=-45, in=-135] (pairn);
\end{scope}

% Phase 3
\begin{scope}[yshift=-3cm]
    \node[llm, right=0.8cm of pair1] (llm6) {\begin{tabular}{@{}c@{}}\raisebox{-0.5pt}{\textcolor{cyan}{\small\faRobot}} \\ \raisebox{-1pt}{\tiny LLM}\end{tabular}};
    \node[conclusion, right=0.8cm of llm6] (conc1) {\faStar\ Conc. 1};
    
    \node[font=\footnotesize] at (6,-0.8) {$\vdots$};
    \node[font=\footnotesize] at (7.5,-0.8) {$\vdots$};
    
    \node[llm, right=0.8cm of pairn] (llm7) {\begin{tabular}{@{}c@{}}\raisebox{-0.5pt}{\textcolor{cyan}{\small\faRobot}} \\ \raisebox{-1pt}{\tiny LLM}\end{tabular}};
    \node[conclusion, right=0.8cm of llm7] (concn) {\faStar\ Conc. n};
    
    % Added dotted connections from paper to conclusion LLMs with arrows and lower curves
    \draw[->, dotted, black, thin] (paper3) to[out=20,in=150] (llm6);
    % \draw[->, dashed, gray, thin] (paper3) to[out=-85,in=150] (llm7);
    
    \draw[arrow] (pair1) -- node[above, font=\tiny] {\begin{tabular}{c} conclusion \\ prompt \end{tabular}} (llm6);
    \draw[arrow] (llm6) -- (conc1);
    \draw[arrow] (pairn) -- node[above, font=\tiny] {\begin{tabular}{c} conclusion \\ prompt \end{tabular}} (llm7);
    \draw[arrow] (llm7) -- (concn);
\end{scope}

% Phase boxes and labels
\node[phase, fit=(paper3)(llm3)(allclaims)] (phase1box) {};
\node[below, font=\footnotesize] at (phase1box.south) {\textbf{Phase 1:} Extract All Claims};

\node[phase, fit=(claim1)(llm4)(evidence1)(pair1)(claimn)(llm5)(evidencen)(pairn)] (phase2box) {};
\node[below, font=\footnotesize] at (phase2box.south) {\textbf{Phase 2:} Extract Evidence \& Form C-E Pairs};

\node[phase, fit=(llm6)(conc1)(llm7)(concn)] (phase3box) {};
\node[below, font=\footnotesize] at (phase3box.south) {\textbf{Phase 3:} Generate Conclusions};

\draw[arrow] (allclaims) -- ($(allclaims)+(0,-1.5)$) -| (claim1);  % Increased vertical gap

\end{tikzpicture}

\caption{Three methods to prompt LLMs to analyze the papers. \textbf{Single-Pass:} Full paper processing with one prompt.
     \textbf{Three-Pass:} Sequential claim $\rightarrow$ evidence $\rightarrow$ conclusion extraction.
     \textbf{One-by-One Pass}: Individual evidence retrieval per claim. }
\label{fig:approaches}
\end{figure*}

\begin{figure*}[!h]
    \centering
    \includegraphics[width=\textwidth]{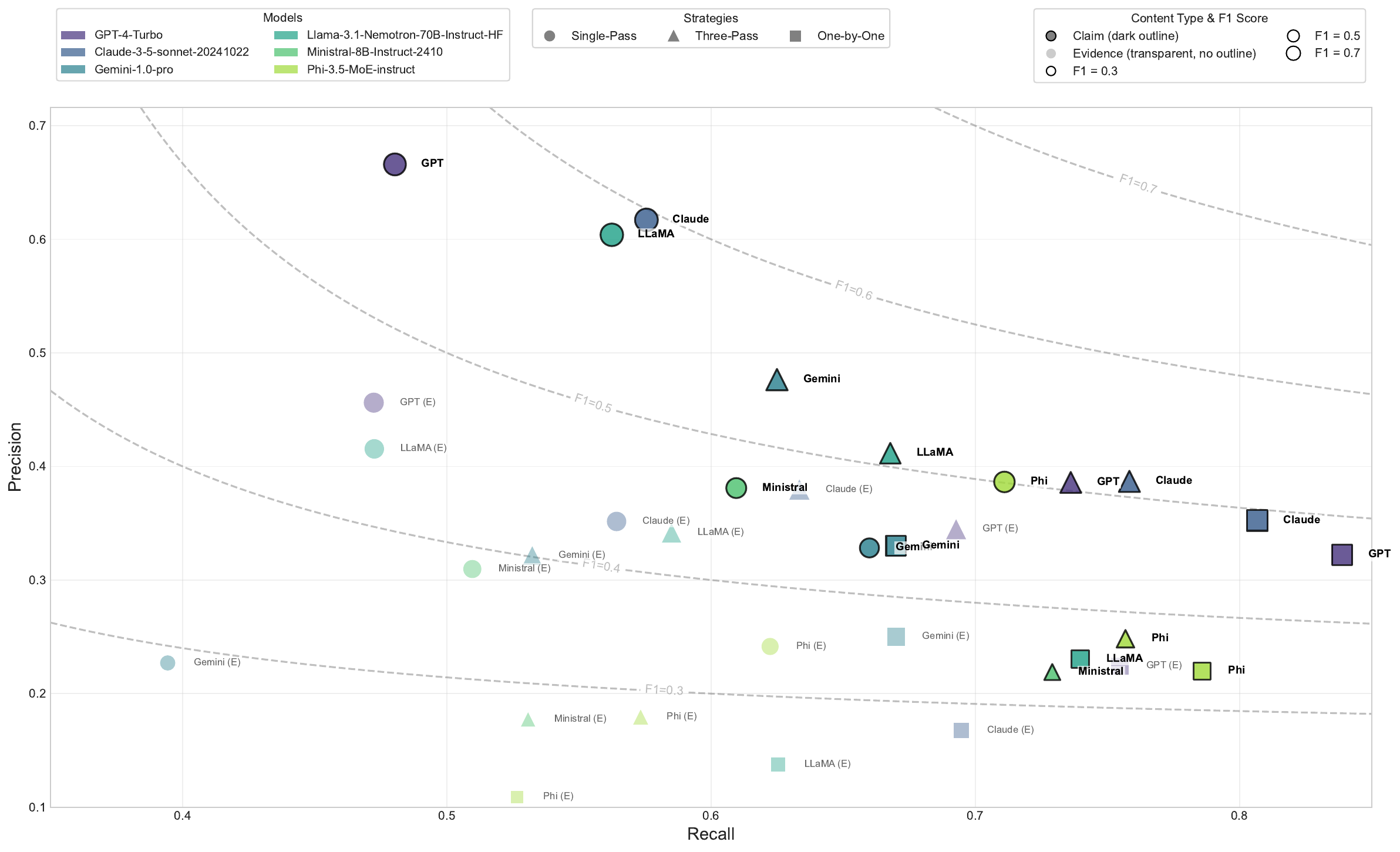}
    \caption{Precision vs. Recall for claim (solid markers) and evidence (transparent markers) identification across models and strategies (shapes: Single-Pass \(\bullet\), Three-Pass \(\blacktriangle\), One-by-One \(\blacksquare\)). Models show higher precision for claims, higher recall for evidence, with most results below \(F_1=0.7\).}

    \label{fig:precision-recall}
\end{figure*}

\subsection{Evaluation Metrics}

\label{Evaluation_metrics}

\newcommand{\TP}{\mathrm{TP}}
\newcommand{\FP}{\mathrm{FP}}
\newcommand{\FN}{\mathrm{FN}}
In this study, we employ four metrics to evaluate the LLM performance: three established metrics in information retrieval, precision, recall, F1-score, and a novel metric, sentence\_gap, to evaluate LLM performance in claim-evidence retrieval tasks and the effectiveness of our various prompting techniques.

\paragraph{Precision (P)} Used to measure the proportion of spans the model predicts that are identified by the annotators, reflecting their effectiveness in responding to precise and carefully structured prompts.
\begin{equation}
    P = \frac{\TP}{\TP + \FP}, 
    \label{eq:precision}
\end{equation}
where $\TP$ (true positive) is the number of correctly retrieved claim/evidence, and $\FP$ (false positive) is the number of retrieved ``claim''/``evidence'' that are not claims/evidences.

\paragraph{Recall (R)} Quantifying the portion of claim/evidence that are retrieved. Recall assesses the ability to capture pertinent data, a measure of the model's responsiveness to exhaustive prompt inquiries 
\begin{equation}
    R = \frac{\TP}{\TP + \FN}, 
    \label{eq:recall}
\end{equation}
where $\FN$ (false negative) is the number of claims/evidences that are incorrectly missed.

\paragraph{F1-score} This is the harmonic mean of P and R. The F1-score provides a balanced measure of accuracy, crucial for evaluating the efficacy of the prompting techniques in eliciting detailed and relevant responses. %\ZZ{Actually this sentence indicates that we should use F1 unless we want to particularly emphasize the precision or recall.} 

\paragraph{sentence\_gap} The sentence\_gap metric measures the distance between a retrieved claim and each of its associated retrieved evidence. It is particularly valuable for evaluating long-range contextual comprehension by quantitatively assessing models' ability to handle textual relationships over extended contexts. This assessment is crucial for complex prompts designed to challenge such comprehension and is instrumental as we explore how increasing LLM context length capabilities enhance performance in realistic scenarios.
\begin{equation}
    \text{sentence\_gap} = \frac{1}{|\mathcal{M}|}\sum_{(p,g)\in\mathcal{M}}\bigl|s(p)-s(g)\bigr|,  \label{eq:sg}
\end{equation}
where $\mathcal{M}$ is the set of matched evidence pairs (using Intersection over Union matching rule). $s(\cdot)$ returns the sentence index of a span inside the document. The sentence\_gap metric is therefore the average absolute sentence-level distance between each predicted claim span $p$ and its evidence span $g$, capturing how far a model must reason across the paper to link claims with supporting evidence.

% \vspace{-0.4cm}

\paragraph{Secondary metrics}
Additionally, we consider secondary metrics that focus on operational aspects of model performance: the time to generate outputs and how each model’s recall changes as input length (token count) increases. These metrics are crucial for understanding efficiency and scalability. They help compare how models manage computational resources and handle large input sizes under varying conditions.

% \vspace{-0.4cm}

\section{Experimental Setup}

% This section details the experimental design of our study, outlining the criteria for selecting the Large Language Models and the various analytical methods implemented to assess their performance on the CLAIM-BENCH dataset.
% \subsection{Model Selection}
\label{Model_selection}
% In this work, we have considered a diverse set of state-of-the-art models for comparison and evaluations, we have considered both open-source models as well as closed-source models, not only that we tried to have diverse range of models with different architectures as well, GPT-4 \citep{openai_gpt-4_2024},
% Gemini-Exp\_1114 \citep{geminiteam2024gemini15unlockingmultimodal}, Claude 3.5 Sonnet \citep{anthropic2025claude35}, Ministral-8B, Phi-3.5-MoE \citep{abdin2024phi3technicalreporthighly} , and LLaMA-70B \citep{wang2025helpsteer2preferencecomplementingratingspreferences}. One of the core reason to pick these models is not only that these are state-of-the-art models but also these are the models that have a context window of 128k or above. 

We evaluate six state‑of‑the‑art LLMs, chosen to span both licensing regimes and architectural families while sharing a $\geq$128K‑token context window. Open‑source include Ministral‑8B \cite{mistral_ministraux_2024}, Phi‑3.5‑MoE \citep{abdin2024phi3technicalreporthighly}, and LLaMA‑70B \citep{wang2025helpsteer2preferencecomplementingratingspreferences} and Closed‑source includes GPT‑4 \citep{openai_gpt-4_2024}, Gemini‑Exp\_1114 \citep{geminiteam2024gemini15unlockingmultimodal}, and Claude 3.5 Sonnet \citep{anthropic2025claude35}.

\subsection{Analysis Methods}
\label{analysis_methods}

As illustrated in \autoref{fig:approaches}, we explore three distinct prompting methods to assess and enhance model performance on claim-evidence identification tasks.

\paragraph{Single-Pass} Initially, we present the models with a research paper, instructing (Appendix \ref{1-pass_prompt}) them to identify claims, evidences, and conclusions in a single comprehensive prompt. 

\paragraph{Three-Pass} Building on the ``divide and conquer'' strategy from prior research, we then deconstruct the task into sequential stages. In the first stage, the model identifies claims using a dedicated prompt. Subsequently, these claims are supplied to the next stage, where separate prompts elicit corresponding evidences. Finally, we combine the identified claims and evidences, using another prompt to extract conclusions (Appendix \ref{3-pass_prompt}).

\begin{figure*}[h!]
    \centering
    \includegraphics[width=0.94\textwidth, height = 0.37\textheight]{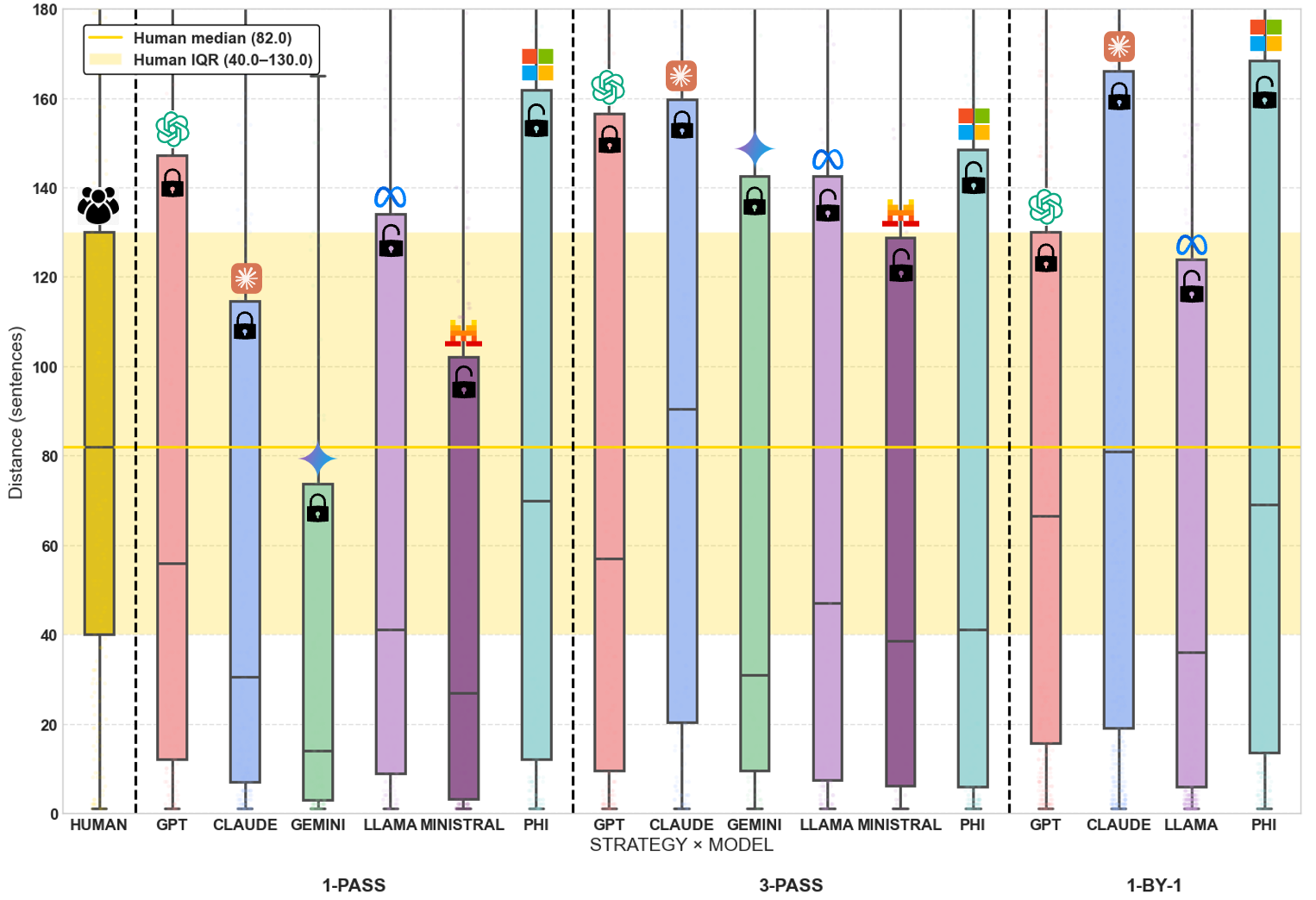}
    \caption{Sentence distance distribution (box plots) between claims and linked evidence vs. Human baseline (\(\mathbf{\textit{leftmost}}\)). LLMs, especially with iterative strategies, link over longer distances than humans, showing capability but potential noise.}

    \label{fig:sent_dist}
\end{figure*}
\paragraph{One-by-One Pass}  We adopt a more granular approach where each claim is processed individually to retrieve evidence. This means for n claims, the model runs 
n times to gather evidence for each, and similarly for conclusions. Although this approach provides detailed analysis, it significantly increases the demand on computational resources and time (Appendix \ref{1_by_1_prompt}). These methods combine careful prompting with our annotated claim–evidence dataset, allowing us to benchmark each model’s extraction accuracy and probe how different prompt strategies improve performance.

\section{Results}

The following section details the experimental results, highlighting comparative model performance and strategic impacts.

\label{Results}

% In this section, we present a multi-faceted evaluation of LLMs using CLAIM-BENCH. Our analysis focuses on three key aspects: (1) how well models balance precision and recall in identifying claims and evidence (\autoref{fig:precision-recall}); (2) how effectively each model links claims to their supporting evidence (\autoref{fig:sent_dist});  and (3) how prompting strategies affect runtime and efficiency (\autoref{fig:time_analysisl}). By examining these dimensions together, we uncover important trade-offs that guide both practical applications and future model improvements.

% \subsection{Claim–Evidence Extraction Performance} \todo{might need a change}

\subsection{Precision vs Recall}
% \ZZ{I feel this paragraph actually discusses one-by-one vs Single-Pass... Perhaps focus more on the P-R patterns, e.g., precision-recall tradeoff? Mention this tradeoff earlier in the paragraph.}
% As shown in \autoref{fig:precision-recall}, the one-by-one mode tends to push models towards high recall, leading them to extract a number of claim–evidence pairs, often with a drop in precision. This effect is particularly noticeable with models like Claude and LLaMA when using this strategy. This is reflected in \autoref{fig:sent_dist}, where these high-recall settings often report very high maximum distances (e.g., LLaMA exceeding 2,200 sentences and Claude showing high variance), indicating that the models are linking claims to evidence from far-flung sections. While such long-range linking is beneficial for capturing subtle, dispersed relationships, it can also introduce more false positives. In contrast, Single-Pass strategies—exemplified by GPT—prioritize a more focused retrieval, yielding moderate max distances (around 658–708 sentences) and consistent variance, suggesting fewer spurious matches at the cost of slightly lower coverage. Ministral displays balanced recall and precision, characterized by relatively consistent and shorter linking distances.

As shown in \autoref{fig:precision-recall}, models exhibit a clear precision-recall trade-off: settings that achieve higher recall often incur reduced precision. For instance, Claude and LLaMA achieve high recall but at the cost of extracting numerous false positives, which is evident from their large maximum linking distances (\autoref{fig:Sentence_distance_analysis}), exceeding 2,200 sentences in some cases. Although valuable, such long-range links raise the risk of false claim–evidence pairs. Conversely, models like GPT prioritize precision, maintaining moderate linking distances (around 658–708 sentences) with fewer spurious matches, though this approach slightly limits recall. Ministral offers a balanced precision-recall profile, characterized by consistent, shorter linking distances.

% \subsection{Open Source vs Closed Source Models}
% \ZZ{I don't think it's necessary to use two subsections to describe model differences. Perhaps merge ``Precision and Recall'' into 5.1 Precision and Recall, and merge ``Coverage and Distance'' into 5.6 Impact of Token Length.}
Comparing the precision-recall tradeoff trends between open- and closed-source models, we see that closed-source models balance precision and recall better. Overall, GPT often balances high precision and moderate recall; Claude achieves higher recall rates but exhibits noticeable trade-offs in precision. Gemini remains stable across strategies. Among open-source models, LLaMA came close to matching closed-source recall but with some outliers, also shows variability in precision; Ministral is moderate in both coverage \& precision; Phi exhibits the widest swings, at times matching larger models but also dropping in accuracy.

\begin{figure*}[ht]
    \centering
    \includegraphics[width=0.95\textwidth]{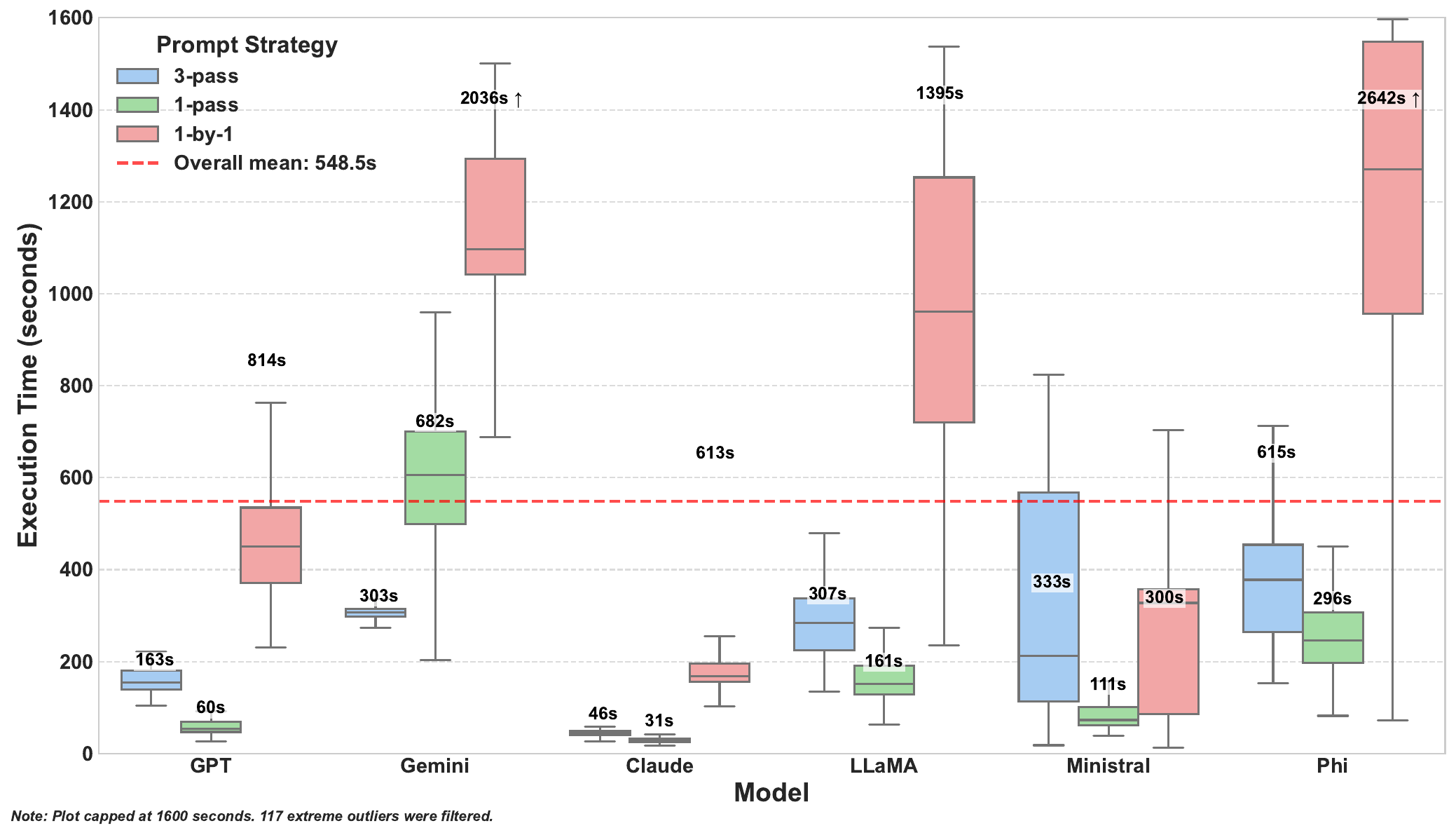}
    \caption{Execution time comparison (box plots): Single-Pass (\textcolor{strategycolor_1pass}{\(\blacksquare\)}) is fastest, One-by-One (\textcolor{strategycolor_1by1}{\(\blacksquare\)}) is slowest. Models vary greatly in speed (e.g., Claude consistently fast; LLaMA/Phi often requiring >1000s).}

    \label{fig:time_analysis}
\end{figure*}
\subsection{ Smaller vs Larger Models}

Larger models, such as GPT-4-Turbo, Claude, Gemini, and LLaMA, generally exhibit strong recall in identifying claims, with GPT-4-Turbo achieving high precision (0.68) and recall (0.81), demonstrating effective balance at different strategies. Claude also shows strong recall (0.83), albeit with a moderate precision drop (0.61). Also, LLaMA achieves similar recall (0.76) but comparative precision (0.60), indicating a tendency to identify extensive and highly precise connections, considering the best cases of each model.

Smaller models, such as Ministral and Phi, typically exhibit lower recall and precision. Ministral shows modest recall (0.60) with precision around 0.38, reflecting a conservative approach to claim-evidence linking. Phi demonstrates similar precision (approximately 0.39) but notably higher recall (around 0.7) in the best cases. These observations highlight a clear trade-off: larger models generally identify broader and more nuanced claim–evidence relationships but often at the cost of precision, whereas smaller models maintain more consistent precision with significantly reduced recall. In both the cases similar pattern holds in evidence extraction as well.

\subsection{Claims vs Evidence Extraction}

\begin{table}[ht]
% \scriptsize
\small
\centering
\setlength{\tabcolsep}{6pt}  % tighten columns
\resizebox{\linewidth}{!}{
\begin{tabular}{lcccccc}
\toprule
\multirow{2}{*}{\textbf{Model}} & \multicolumn{3}{c}{\textbf{Best C Performances}} & \multicolumn{3}{c}{\textbf{Best E Performances}} \\
& F1 & P & R & F1 & P & R \\
\midrule
GPT-4-Turbo        & 0.56 &0.66 & 0.57 & \textbf{0.47} & \textbf{0.34} & \textbf{0.69}   \\
Claude 3.5         & \textbf{0.59} &\textbf{0.62} & \textbf{0.60} & 0.42 & 0.33  & 0.66  \\
Gemini-Exp\_1114   & 0.54 & 0.48 & 0.64 & 0.40 & 0.30 & 0.52 \\
LLaMA-70B          & 0.58 & 0.60 & 0.56 & 0.45 & 0.42 & 0.49 \\
Ministral-8B         & 0.48 & 0.39 & 0.61 & 0.39 & 0.31 & 0.52 \\
Phi-3.5-MoE        & 0.50 &0.40 & 0.72 & 0.35 & 0.25 & 0.63 \\
\bottomrule
\end{tabular}}
\caption{The highest performance (across all strategies) for Claim (C) and Evidence (E) extraction; ``P@R'' denotes precision at the corresponding recall.}
\label{claim_vs_evi_table}
\end{table}

Analyzing claim versus evidence extraction separately reveals distinct performances among LLMs (see Table \ref{claim_vs_evi_table}). Across all models, precision is consistently higher for claims than for evidence, indicating the models more readily detect explicit claims compared to the contextually dispersed evidence. Also, the evidence extraction of all models yields higher recall than precision. 
In addition to the common trends, the models exhibit distinct patterns. For instance, Claude and LLaMA exhibit high recall in evidence extraction but with substantial variability in linking distances (Claude: mean gap of 119.4 sentences, variance of 33,674; LLaMA: mean 95.1 sentences, variance of 34,207), suggesting increased noise and inconsistent performance. Conversely, Ministral maintains lower linking distances (mean 75.9 sentences) with minimal variance, signifying a more cautious and controlled approach.
% \ZZ{When possible, make the first sentence of 5.1 to 5.7 an informative summary sentence of that subsection. For example, in 5.4, don't just say ``the performances differ''. Instead, say ``C performance is higher than E''}
% Analyzing claim versus evidence extraction separately highlights significant performance differences among LLMs. Models typically demonstrate higher precision in claim extraction than in evidence extraction. GPT-4 achieves around 0.7 precision for claims but drops to about 0.35 for evidence, while Claude similarly shows claim precision around 0.65 versus about 0.3 for evidence. This discrepancy between GPT4 and Claude indicates models more readily detect explicit claims than they do dispersed, context-dependent evidence.

% Evidence extraction generally shows higher recall but lower precision \ZZ{than F1?}. Claude and LLaMA demonstrate strong recall (0.6–0.8) but experience noise, seen in higher sentence gaps (Claude: 119.4, LLaMA: 95.1). Ministral maintains conservative precision ($\sim$0.3) and lower sentence gaps (75.9 in three-pass with 38 pairs), suggesting a cautious but precise approach. Phi exhibits variability, with a mean distance of 105.9 and high variance (13,188.2) in one-by-one strategy, highlighting inconsistency in evidence linking.

\subsection{Impact of Strategy}
The Single-pass strategy is highly efficient but has limited coverage, e.g., GPT-4 produces 152 pairs with a 98.5 average sentence\_gap, while Ministral generates 166 pairs (average gap: 64.2). 
Meanwhile, the Three-pass strategy enhances recall and coverage at moderate computational cost. Claude yields 174 pairs (average gap: 122.2), and Phi captures 279 pairs, albeit with significant variance (11,490.2) in sentence\_gap. 
Finally, the One-by-One strategy maximizes recall but increases computational demand significantly. Claude and LLaMA produce the highest counts (639 and 659 pairs, respectively), with substantial gaps (Claude: 119.4, LLaMA: 95.1) and high variance (Claude: 33,673.9, LLaMA: 34,207.0). Phi also achieves substantial coverage (347 pairs) with notable variance (13,188.2).

% \vspace{-0.2cm}
\subsection{Impact of Token Length on Recall }

We observed how the documents' token length affected the models' recall performances. In long documents, we expected performance drops, but these observed drops are tied to the prompting strategy. 
With the Single-pass strategy, the recall performances dropped as the document length increased.
With the iterative prompting strategies (Three-pass or One-by-One), the performance drops are less significant, indicating that the iterative prompting imposes less ``processing load'' onto the LLMs.
Additionally, the recall drops differ by the sizes of the models. Relatively smaller models (LLaMA 70B and Ministral 8B) showed more notable declines, especially with Single-pass, whereas the larger models (Claude and GPT-4) maintained relatively high recalls, underlining the advantage of their long context capabilities. Additional details in Appendix \ref{token_length}.

Claude and LLaMA frequently produce the highest pair counts (up to 639 and 659), reflecting broad coverage. This can coincide with their large context window sizes—helpful for capturing distant relationships—yet also introduces potential noise. GPT and Gemini keep moderate distances, suggesting they discovered fewer links. Ministral remains conservative with fewer pairs with shorter distances, while Phi’s extreme variance indicates inconsistent linking across long contexts. We include the details in Figure \ref{fig:Sentence_distance_analysis} (in Appendix \ref{token_length}).

%% SS: Maybe I will change the sentences then? Ok
% Perhaps let's make the plot later , and add it to the appendix in July, Okay,
% \vspace{-0.2cm}

\subsection{Execution Time Analysis} 
As shown in \autoref{fig:time_analysis}, the execution times differ considerably across models and strategies. GPT is highly efficient in the Single-Pass (under 200s) and relatively moderate in one-by-one approaches ($\sim$500s). Gemini exhibits intermediate execution times across all strategies, notably higher for the three-pass ($\sim$600s). Claude consistently achieves the fastest execution across all strategies, maintaining execution times under 200 seconds. LLaMA shows extensive variability, especially with one-by-one strategies frequently exceeding 1,200 seconds, reflecting significant computational demands. Ministral shows relatively balanced execution times, with three-pass and one-by-one strategies averaging around 600–900 seconds. Phi demonstrates the highest computational intensity, especially in one-by-one strategies, often surpassing 1,200 seconds, highlighting the considerable resource investment required for thorough analyses. The execution times recorded for Gemini exhibit some variability, which may partially stem from fluctuations in API response latency during our experiments, combined with the necessary sleep() intervals implemented for rate limiting. 
% \ZZ{Could you also use another paragraph to comment on the token count? Compared to the time, token count is linked more directly to the cost, so people want to see that as well. At least mention things like ``in each second you are going to burn 1 dollar''.}

\section{Discussion}
The insights from CLAIM-BENCH emphasize critical directions for future research and practical applications leveraging the capabilities of LLMs in scientific claim-evidence reasoning. Improving LLMs' ability to accurately validate claim-evidence pairs could enhance their practical use in designing experiments and generating scientifically valid hypotheses. Furthermore, improved claim identification and validation methods provide a foundation for developing sophisticated claim quality scoring tools that can greatly enhance peer-review processes. The capability to systematically link and integrate evidence across multiple scientific papers could lead to powerful retrieval-augmented laboratory assistants and cross-paper evidence graphs, accelerating knowledge discovery. These advancements would not only strengthen the robustness of scientific validations but also facilitate the creation of more sophisticated scientific QA systems, thus laying foundational benchmarks for future scientific text generation and evaluation methods. This research thus serves as a pivotal foundation for transformative applications in scientific inquiry and discourse.

\section{Conclusion}

Motivated by the limited evaluation in prior literature of LLMs' abilities in scientific reasoning, we introduced CLAIM-BENCH, a novel benchmark specifically designed to evaluate LLMs' capabilities in identifying and validating claim-evidence relationships within scientific texts. 
We systematically explored diverse LLM architectures and prompting strategies. 
% \ZZ{Summarize the key findings. Use each sentence to only describe \textit{one} subsection among 5.1 to 5.7.} 
Our results demonstrate significant limitations in LLMs' comprehension, specifically in their precision and recall balance when processing complex scientific documents. Notably, models showed higher precision in extracting explicit claims, whereas extracting dispersed evidence proved challenging, yielding higher recall but lower precision and increased sentence gaps. Moreover, our comparative analysis across 3 strategies revealed substantial trade-offs between computational efficiency, precision, and coverage. Closed-source models generally displayed more stable performances, while open-source models offered broad yet inconsistent coverage. 
CLAIM-BENCH provides a framework for the assessment of LLMs in complex scientific contexts, and our study provides useful material and insights for continuing the advancement in LLMs' high-level comprehension and scientific reasoning capabilities.

\section{Limitations}

While CLAIM-BENCH provides comprehensive insights into the capabilities of LLMs in scientific claim-evidence reasoning. Despite these insights, CLAIM-BENCH has several limitations worth noting. First, the benchmark primarily focuses on recent papers from select domains, which are after the LLMs' knowledge cutoff but might limit the generalizability. Second, the evaluation relies on existing LLM architectures. While we leave the exploration of the impact of model architecture development to future works, CLAIM-BENCH could be a useful material that supports future projects that develop novel LLM architectures that have enhanced long-context language understanding capabilities and scientific reasoning capabilities.

\bibliography{latex/acl_latex}

\begin{thebibliography}{34}
\providecommand{\natexlab}[1]{#1}

\bibitem[{Abdin et~al.(2024)Abdin, Aneja, and et~al.}]{abdin2024phi3technicalreporthighly}
Marah Abdin, Jyoti Aneja, and Hany~Awadalla et~al. 2024.
\newblock \href {https://arxiv.org/abs/2404.14219} {Phi-3 technical report: A highly capable language model locally on your phone}.
\newblock \emph{Preprint}, arXiv:2404.14219.

\bibitem[{Agarwal et~al.(2025)Agarwal, Sahu, Puri, Laradji, Dvijotham, Stanley, Charlin, and Pal}]{agarwal_litllm:_2025}
Shubham Agarwal, Gaurav Sahu, Abhay Puri, Issam~H. Laradji, Krishnamurthy~DJ Dvijotham, Jason Stanley, Laurent Charlin, and Christopher Pal. 2025.
\newblock \href {https://doi.org/10.48550/arXiv.2402.01788} {{LitLLM}: {A} {Toolkit} for {Scientific} {Literature} {Review}}.
\newblock \emph{arXiv preprint}.
\newblock ArXiv:2402.01788.

\bibitem[{Anthropic(2025)}]{anthropic2025claude35}
Anthropic. 2025.
\newblock \href {https://www-cdn.anthropic.com/fed9cc193a14b84131812372d8d5857f8f304c52/Model_Card_Claude_3_Addendum.pdf} {Claude 3.5 sonnet model card addendum}.
\newblock PDF file. Accessed 12 Apr. 2025.

\bibitem[{Chang et~al.(2024)Chang, Li, Jia, Wang, Huang, Wang, Huang, and Liu}]{chang_what_2024}
Zhiyuan Chang, Mingyang Li, Xiaojun Jia, Junjie Wang, Yuekai Huang, Qing Wang, Yihao Huang, and Yang Liu. 2024.
\newblock \href {https://doi.org/10.48550/arXiv.2412.12632} {What {External} {Knowledge} is {Preferred} by {LLMs}? {Characterizing} and {Exploring} {Chain} of {Evidence} in {Imperfect} {Context}}.
\newblock \emph{arXiv preprint}.
\newblock ArXiv:2412.12632.

\bibitem[{Checco et~al.(2021)Checco, Bracciale, Loreti, Pinfield, and Bianchi}]{checco_ai-assisted_2021}
Alessandro Checco, Lorenzo Bracciale, Pierpaolo Loreti, Stephen Pinfield, and Giuseppe Bianchi. 2021.
\newblock \href {https://doi.org/10.1057/s41599-020-00703-8} {{AI}-assisted peer review}.
\newblock \emph{Humanities and Social Sciences Communications}, 8(1):1--11.

\bibitem[{Chen et~al.(2025)Chen, Chen, Ning, Zhang, Wang, Yu, Li, Liao, Wei, Lu, Dey, Xue, Baker, Burns, Adu-Ampratwum, Huang, Ning, Gao, Su, and Sun}]{chen_scienceagentbench:_2025}
Ziru Chen, Shijie Chen, Yuting Ning, Qianheng Zhang, Boshi Wang, Botao Yu, Yifei Li, Zeyi Liao, Chen Wei, Zitong Lu, Vishal Dey, Mingyi Xue, Frazier~N. Baker, Benjamin Burns, Daniel Adu-Ampratwum, Xuhui Huang, Xia Ning, Song Gao, Yu~Su, and Huan Sun. 2025.
\newblock \href {https://openreview.net/forum?id=6z4YKr0GK6} {{ScienceAgentBench}: {Toward} {Rigorous} {Assessment} of {Language} {Agents} for {Data}-{Driven} {Scientific} {Discovery}}.
\newblock In \emph{The {Thirteenth} {International} {Conference} on {Learning} {Representations}}.

\bibitem[{Chernyshev et~al.(2025)Chernyshev, Polshkov, Artemova, Myasnikov, Stepanov, Miasnikov, and Tilga}]{chernyshev_u-math:_2025}
Konstantin Chernyshev, Vitaliy Polshkov, Ekaterina Artemova, Alex Myasnikov, Vlad Stepanov, Alexei Miasnikov, and Sergei Tilga. 2025.
\newblock \href {https://doi.org/10.48550/arXiv.2412.03205} {U-{MATH}: {A} {University}-{Level} {Benchmark} for {Evaluating} {Mathematical} {Skills} in {LLMs}}.
\newblock \emph{arXiv preprint}.
\newblock ArXiv:2412.03205.

\bibitem[{Costa-jussà et~al.(2024{\natexlab{a}})Costa-jussà, Andrews, Meglioli, Chen, Chuang, Dale, Ropers, Mourachko, Sánchez, Schwenk, Tran, Turkatenko, and Wood}]{costa-jussa_lcfo:_2024}
Marta~R. Costa-jussà, Pierre Andrews, Mariano~Coria Meglioli, Joy Chen, Joe Chuang, David Dale, Christophe Ropers, Alexandre Mourachko, Eduardo Sánchez, Holger Schwenk, Tuan Tran, Arina Turkatenko, and Carleigh Wood. 2024{\natexlab{a}}.
\newblock \href {https://doi.org/10.48550/arXiv.2412.08268} {{LCFO}: {Long} {Context} and {Long} {Form} {Output} {Dataset} and {Benchmarking}}.
\newblock \emph{arXiv preprint}.
\newblock ArXiv:2412.08268.

\bibitem[{Costa-jussà et~al.(2024{\natexlab{b}})Costa-jussà, Chen, Adebara, Chuang, Ropers, and Sánchez}]{costa-jussa_y-nq:_2024}
Marta~R. Costa-jussà, Joy Chen, Ifeoluwanimi Adebara, Joe Chuang, Christophe Ropers, and Eduardo Sánchez. 2024{\natexlab{b}}.
\newblock \href {https://doi.org/10.48550/arXiv.2412.08279} {Y-{NQ}: {English}-{Yorùbá} {Evaluation} dataset for {Open}-{Book} {Reading} {Comprehension} and {Text} {Generation}}.
\newblock \emph{arXiv preprint}.
\newblock ArXiv:2412.08279.

\bibitem[{Dingjie et~al.(2024)Dingjie, Chen, Chen, Yu, Wan, and Wang}]{dingjie_milebench:_2024}
Song Dingjie, Shunian Chen, Guiming~Hardy Chen, Fei Yu, Xiang Wan, and Benyou Wang. 2024.
\newblock \href {https://openreview.net/forum?id=Uhwze2LEwq} {{MileBench}: {Benchmarking} {MLLMs} in {Long} {Context}}.
\newblock In \emph{First {Conference} on {Language} {Modeling}}.

\bibitem[{{Gemini Team}(2024)}]{geminiteam2024gemini15unlockingmultimodal}
{Gemini Team}. 2024.
\newblock \href {https://arxiv.org/abs/2403.05530} {Gemini 1.5: Unlocking multimodal understanding across millions of tokens of context}.
\newblock \emph{Preprint}, arXiv:2403.05530.

\bibitem[{Godbole et~al.(2024)Godbole, George, and Shandilya}]{godbole_leveraging_2024}
Aditi Godbole, Jabin~Geevarghese George, and Smita Shandilya. 2024.
\newblock \href {https://doi.org/10.48550/arXiv.2409.18454} {Leveraging {Long}-{Context} {Large} {Language} {Models} for {Multi}-{Document} {Understanding} and {Summarization} in {Enterprise} {Applications}}.
\newblock \emph{arXiv preprint}.
\newblock ArXiv:2409.18454.

\bibitem[{Hong et~al.(2024)Hong, Lin, Liu, Liu, Wu, Zhang, Wei, Li, Chen, Zhang, Wang, Zhang, Zhang, Yang, Zhuge, Guo, Zhou, Tao, Tang, Lu, Zheng, Liang, Fei, Cheng, Gou, Xu, and Wu}]{hong_data_2024}
Sirui Hong, Yizhang Lin, Bang Liu, Bangbang Liu, Binhao Wu, Ceyao Zhang, Chenxing Wei, Danyang Li, Jiaqi Chen, Jiayi Zhang, Jinlin Wang, Li~Zhang, Lingyao Zhang, Min Yang, Mingchen Zhuge, Taicheng Guo, Tuo Zhou, Wei Tao, Xiangru Tang, Xiangtao Lu, Xiawu Zheng, Xinbing Liang, Yaying Fei, Yuheng Cheng, Zhibin Gou, Zongze Xu, and Chenglin Wu. 2024.
\newblock \href {https://doi.org/10.48550/arXiv.2402.18679} {Data {Interpreter}: {An} {LLM} {Agent} {For} {Data} {Science}}.
\newblock \emph{arXiv preprint}.
\newblock ArXiv:2402.18679.

\bibitem[{Jin et~al.(2024)Jin, Zhao, Wang, Chen, Zhu, Xiao, and Wang}]{jin_agentreview:_2024}
Yiqiao Jin, Qinlin Zhao, Yiyang Wang, Hao Chen, Kaijie Zhu, Yijia Xiao, and Jindong Wang. 2024.
\newblock \href {https://doi.org/10.48550/arXiv.2406.12708} {{AgentReview}: {Exploring} {Peer} {Review} {Dynamics} with {LLM} {Agents}}.
\newblock \emph{arXiv preprint}.
\newblock ArXiv:2406.12708.

\bibitem[{Leng et~al.(2024)Leng, Wang, and Yuan}]{leng_llm-assisted_2024}
Yan Leng, Hao Wang, and Yuan Yuan. 2024.
\newblock \href {https://papers.ssrn.com/abstract=4948029} {Llm-{Assisted} {Hypothesis} {Generation} and {Graph}-{Based} {Evaluation}}.

\bibitem[{LI et~al.(2025)LI, Jiang, Wu, Luo, Ahn, Zhang, Abdi, Li, Gao, Yang, and Qiu}]{li_scbench:_2025}
YUCHENG LI, Huiqiang Jiang, Qianhui Wu, Xufang Luo, Surin Ahn, Chengruidong Zhang, Amir~H. Abdi, Dongsheng Li, Jianfeng Gao, Yuqing Yang, and Lili Qiu. 2025.
\newblock \href {https://openreview.net/forum?id=gkUyYcY1W9} {{SCBench}: {A} {KV} {Cache}-{Centric} {Analysis} of {Long}-{Context} {Methods}}.
\newblock In \emph{The {Thirteenth} {International} {Conference} on {Learning} {Representations}}.

\bibitem[{Li et~al.(2025)Li, Chen, Liu, Yu, and Wen}]{li_chatcite:_2025}
Yutong Li, Lu~Chen, Aiwei Liu, Kai Yu, and Lijie Wen. 2025.
\newblock \href {https://aclanthology.org/2025.coling-main.244/} {{ChatCite}: {LLM} {Agent} with {Human} {Workflow} {Guidance} for {Comparative} {Literature} {Summary}}.
\newblock In \emph{Proceedings of the 31st {International} {Conference} on {Computational} {Linguistics}}, pages 3613--3630, Abu Dhabi, UAE. Association for Computational Linguistics.

\bibitem[{Lissack and Meagher(2024)}]{lissack_ethical_2024}
Michael Lissack and Brenden Meagher. 2024.
\newblock \href {https://doi.org/10.2139/ssrn.4950138} {Ethical {Use} of {Large} {Language} {Models} in {Academic} {Research} and {Writing}: {A} {How}-{To}}.

\bibitem[{Liu et~al.(2025)Liu, Gwizdka, and Lease}]{liu_exploring_2025}
Houjiang Liu, Jacek Gwizdka, and Matthew Lease. 2025.
\newblock \href {https://doi.org/10.48550/arXiv.2412.08185} {Exploring {Multidimensional} {Checkworthiness}: {Designing} {AI}-assisted {Claim} {Prioritization} for {Human} {Fact}-checkers}.
\newblock \emph{arXiv preprint}.
\newblock ArXiv:2412.08185.

\bibitem[{Liu and Shah(2023)}]{liu_reviewergpt?_2023}
Ryan Liu and Nihar~B. Shah. 2023.
\newblock \href {https://doi.org/10.48550/arXiv.2306.00622} {{ReviewerGPT}? {An} {Exploratory} {Study} on {Using} {Large} {Language} {Models} for {Paper} {Reviewing}}.
\newblock \emph{arXiv preprint}.
\newblock ArXiv:2306.00622.

\bibitem[{Liu et~al.(2024)Liu, Liu, Zhu, Lei, Yang, Zhang, Li, and Liu}]{liu_aigs:_2024}
Zijun Liu, Kaiming Liu, Yiqi Zhu, Xuanyu Lei, Zonghan Yang, Zhenhe Zhang, Peng Li, and Yang Liu. 2024.
\newblock \href {https://doi.org/10.48550/arXiv.2411.11910} {{AIGS}: {Generating} {Science} from {AI}-{Powered} {Automated} {Falsification}}.
\newblock \emph{arXiv preprint}.
\newblock ArXiv:2411.11910.

\bibitem[{Lu et~al.(2024)Lu, Lu, Lange, Foerster, Clune, and Ha}]{lu_ai_2024}
Chris Lu, Cong Lu, Robert~Tjarko Lange, Jakob Foerster, Jeff Clune, and David Ha. 2024.
\newblock \href {https://doi.org/10.48550/arXiv.2408.06292} {The {AI} {Scientist}: {Towards} {Fully} {Automated} {Open}-{Ended} {Scientific} {Discovery}}.
\newblock \emph{arXiv preprint}.
\newblock ArXiv:2408.06292.

\bibitem[{Luo et~al.(2025)Luo, Rechardt, Sun, Nejad, Yáñez, Yilmaz, Lee, Cohen, Borghesani, Pashkov, Marinazzo, Nicholas, Salatiello, Sucholutsky, Minervini, Razavi, Rocca, Yusifov, Okalova, Gu, Ferianc, Khona, Patil, Lee, Mata, Myers, Bizley, Musslick, Bilgin, Niso, Ales, Gaebler, Ratan~Murty, Loued-Khenissi, Behler, Hall, Dafflon, Bao, and Love}]{luo_large_2025}
Xiaoliang Luo, Akilles Rechardt, Guangzhi Sun, Kevin~K. Nejad, Felipe Yáñez, Bati Yilmaz, Kangjoo Lee, Alexandra~O. Cohen, Valentina Borghesani, Anton Pashkov, Daniele Marinazzo, Jonathan Nicholas, Alessandro Salatiello, Ilia Sucholutsky, Pasquale Minervini, Sepehr Razavi, Roberta Rocca, Elkhan Yusifov, Tereza Okalova, Nianlong Gu, Martin Ferianc, Mikail Khona, Kaustubh~R. Patil, Pui-Shee Lee, Rui Mata, Nicholas~E. Myers, Jennifer~K. Bizley, Sebastian Musslick, Isil~Poyraz Bilgin, Guiomar Niso, Justin~M. Ales, Michael Gaebler, N.~Apurva Ratan~Murty, Leyla Loued-Khenissi, Anna Behler, Chloe~M. Hall, Jessica Dafflon, Sherry~Dongqi Bao, and Bradley~C. Love. 2025.
\newblock \href {https://doi.org/10.1038/s41562-024-02046-9} {Large language models surpass human experts in predicting neuroscience results}.
\newblock \emph{Nature Human Behaviour}, 9(2):305--315.

\bibitem[{Ma et~al.(2024)Ma, Zang, Chen, Chen, Jiao, Li, Lu, Liu, Ma, Dong, Zhang, Pan, Jiang, Wang, Cao, and Sun}]{ma_mmlongbench-doc:_2024}
Yubo Ma, Yuhang Zang, Liangyu Chen, Meiqi Chen, Yizhu Jiao, Xinze Li, Xinyuan Lu, Ziyu Liu, Yan Ma, Xiaoyi Dong, Pan Zhang, Liangming Pan, Yu-Gang Jiang, Jiaqi Wang, Yixin Cao, and Aixin Sun. 2024.
\newblock \href {https://openreview.net/forum?id=loJM1acwzf} {{MMLONGBENCH}-{DOC}: {Benchmarking} {Long}-context {Document} {Understanding} with {Visualizations}}.
\newblock In \emph{The {Thirty}-eight {Conference} on {Neural} {Information} {Processing} {Systems} {Datasets} and {Benchmarks} {Track}}.

\bibitem[{{Mistral AI}(2024)}]{mistral_ministraux_2024}
{Mistral AI}. 2024.
\newblock Un {Ministral}, des {Ministraux}: Introducing the world’s best edge models.
\newblock \url{https://mistral.ai/news/ministraux}.
\newblock Accessed 19 May 2025.

\bibitem[{Ni et~al.(2024)Ni, Cai, Wei, Wang, Yin, and Li}]{ni_xl$^2$bench:_2024}
Xuanfan Ni, Hengyi Cai, Xiaochi Wei, Shuaiqiang Wang, Dawei Yin, and Piji Li. 2024.
\newblock \href {https://doi.org/10.48550/arXiv.2404.05446} {{XL}\${\textasciicircum}2\${Bench}: {A} {Benchmark} for {Extremely} {Long} {Context} {Understanding} with {Long}-range {Dependencies}}.
\newblock \emph{arXiv preprint}.
\newblock ArXiv:2404.05446.

\bibitem[{{OpenAI}(2024)}]{openai_gpt-4_2024}
{OpenAI}. 2024.
\newblock \href {https://doi.org/10.48550/arXiv.2303.08774} {{GPT}-4 {Technical} {Report}}.
\newblock \emph{arXiv preprint}.
\newblock ArXiv:2303.08774.

\bibitem[{Su et~al.(2025)Su, Chen, Tang, Yin, Zheng, Li, Qi, Wu, Li, Ouyang, Torr, Zhou, and Dong}]{su_many_2025}
Haoyang Su, Renqi Chen, Shixiang Tang, Zhenfei Yin, Xinzhe Zheng, Jinzhe Li, Biqing Qi, Qi~Wu, Hui Li, Wanli Ouyang, Philip Torr, Bowen Zhou, and Nanqing Dong. 2025.
\newblock \href {https://doi.org/10.48550/arXiv.2410.09403} {Many {Heads} {Are} {Better} {Than} {One}: {Improved} {Scientific} {Idea} {Generation} by {A} {LLM}-{Based} {Multi}-{Agent} {System}}.
\newblock \emph{arXiv preprint}.
\newblock ArXiv:2410.09403.

\bibitem[{Sun et~al.(2024{\natexlab{a}})Sun, Chan, Chang, and Dow}]{sun_reviewflow:_2024}
Lu~Sun, Aaron Chan, Yun~Seo Chang, and Steven~P. Dow. 2024{\natexlab{a}}.
\newblock \href {https://doi.org/10.1145/3640543.3645159} {{ReviewFlow}: {Intelligent} {Scaffolding} to {Support} {Academic} {Peer} {Reviewing}}.
\newblock In \emph{Proceedings of the 29th {International} {Conference} on {Intelligent} {User} {Interfaces}}, pages 120--137, Greenville SC USA. ACM.

\bibitem[{Sun et~al.(2024{\natexlab{b}})Sun, Tao, Hu, and Dow}]{sun_metawriter:_2024}
Lu~Sun, Stone Tao, Junjie Hu, and Steven~P. Dow. 2024{\natexlab{b}}.
\newblock \href {https://doi.org/10.1145/3637371} {{MetaWriter}: {Exploring} the {Potential} and {Perils} of {AI} {Writing} {Support} in {Scientific} {Peer} {Review}}.
\newblock \emph{Proceedings of the ACM on Human-Computer Interaction}, 8(CSCW1):1--32.

\bibitem[{Vladika and Matthes(2023)}]{vladika_scientific_2023}
Juraj Vladika and Florian Matthes. 2023.
\newblock \href {https://doi.org/10.48550/arXiv.2305.16859} {Scientific {Fact}-{Checking}: {A} {Survey} of {Resources} and {Approaches}}.
\newblock \emph{arXiv preprint}.
\newblock ArXiv:2305.16859.

\bibitem[{Wang et~al.(2025)Wang, Bukharin, Delalleau, Egert, Shen, Zeng, Kuchaiev, and Dong}]{wang2025helpsteer2preferencecomplementingratingspreferences}
Zhilin Wang, Alexander Bukharin, Olivier Delalleau, Daniel Egert, Gerald Shen, Jiaqi Zeng, Oleksii Kuchaiev, and Yi~Dong. 2025.
\newblock \href {https://arxiv.org/abs/2410.01257} {Helpsteer2-preference: Complementing ratings with preferences}.
\newblock \emph{Preprint}, arXiv:2410.01257.

\bibitem[{Weng et~al.(2025)Weng, Zhu, Bao, Zhang, Wang, Zhang, and Yang}]{weng_cycleresearcher:_2025}
Yixuan Weng, Minjun Zhu, Guangsheng Bao, Hongbo Zhang, Jindong Wang, Yue Zhang, and Linyi Yang. 2025.
\newblock \href {https://openreview.net/forum?id=bjcsVLoHYs} {{CycleResearcher}: {Improving} {Automated} {Research} via {Automated} {Review}}.
\newblock In \emph{The {Thirteenth} {International} {Conference} on {Learning} {Representations}}.

\bibitem[{Wu et~al.(2025)Wu, Hee, Hu, and Lee}]{wu_longgenbench:_2025}
Yuhao Wu, Ming~Shan Hee, Zhiqiang Hu, and Roy Ka-Wei Lee. 2025.
\newblock \href {https://openreview.net/forum?id=3A71qNKWAS} {{LongGenBench}: {Benchmarking} {Long}-{Form} {Generation} in {Long} {Context} {LLMs}}.
\newblock In \emph{The {Thirteenth} {International} {Conference} on {Learning} {Representations}}.

\end{thebibliography}

\appendix

\label{sec:appendix}

\section{Prompt Templates}
\onecolumn

\subsection{Single-Pass Prompt}
\label{1-pass_prompt}

\lstset{
    basicstyle=\small\ttfamily,
    columns=flexible,
    breaklines=true,
    postbreak=\mbox{\textcolor{red}{$\hookrightarrow$\space}},
    upquote=true,
    frame=single
}

\begin{tcolorbox}[
  title=Comprehensive Evaluation Prompt, 
  colback=white, 
  colframe=blue!75!black, 
  fonttitle=\bfseries, 
  breakable, 
  enhanced, % ensures enhanced features are active
  sharp corners
]
\textbf{Analyze the research paper and provide a comprehensive evaluation following these guidelines:}

\begin{enumerate}
    \item Identify ALL claims in the paper where each claim:
    \begin{itemize}
        \item Makes a specific, verifiable assertion
        \item Is supported by concrete evidence
        \item Represents findings, contributions, or methodological advantages
        \item Can be from any section except abstract
    \end{itemize}

    \item For each identified claim:
    \begin{itemize}
        \item Extract ALL supporting or contradicting evidence (experimental results, data, or methodology)
        \item Evaluate the evidence strength and limitations
        \item Assess how well conclusions align with evidence
    \end{itemize}
\end{enumerate}

Return ONLY the following JSON structure:
\begin{lstlisting}
{
    "analysis": [
        {
            "claim_id": number,
            "claim": {
                "text": "statement of the claim",
                "type": "methodology/result/contribution/performance",
                "location": "section/paragraph",
                "exact_quote": "verbatim text from paper"
            },
            "evidence": [
                {
                    "evidence_text": "specific experimental result/data",
                    "strength": "strong/moderate/weak",
                    "limitations": "specific limitations",
                    "location": "section/paragraph",
                    "exact_quote": "verbatim text from paper"
                }
            ],
            "evaluation": {
                "conclusion_justified": true/false,
                "robustness": "high/medium/low",
                "justification": "explanation of evidence-conclusion alignment",
                "key_limitations": "critical limitations affecting validity",
                "confidence_level": "high/medium/low"
            }
        }
    ]
}
\end{lstlisting}

\textbf{Ensure:}
\begin{itemize}
    \item ALL substantive claims are captured
    \item Evaluations are objective and well-reasoned
    \item All locations and quotes are precise
    \item Multiple pieces of evidence per claim are included when present
\end{itemize}

\end{tcolorbox}

\subsection{Three-Pass Prompt}
\label{3-pass_prompt}

\begin{tcolorbox}[
  title=Claims Extraction Prompt, 
  colback=white, 
  colframe=blue!75!black, 
  fonttitle=\bfseries, 
  breakable, 
  enhanced, % ensures enhanced features are active
  sharp corners
]
\textit{Paper text: \{text\}}

\textbf{Task:} Identify all statements in the text that meet the following criteria for a claim:
\begin{enumerate}
    \item Makes a specific, testable assertion about results, methods, or contributions.
    \item Represents a novel finding, improvement, or advancement.
    \item Presents a clear position or conclusion.
\end{enumerate}

\textbf{Requirements:}
\begin{itemize}
    \item Include both major and minor claims.
    \item Don't miss any claims.
    \item Present each claim as a separate item.
\end{itemize}

\textbf{Return ONLY the following JSON structure:}
\begin{lstlisting}
{
    "claims": [
        {
            "claim_id": 1,
            "claim_text": "statement of the claim",
            "location": "section/paragraph where this claim appears",
            "claim_type": "Nature of the claim",
            "exact_quote": "complete verbatim text containing the claim"
        }
    ]
}
\end{lstlisting}
\end{tcolorbox}

% \newpage

\begin{tcolorbox}[
  title=Evidence Identification Prompt, 
  colback=white, 
  colframe=red!75!black, 
  fonttitle=\bfseries, 
  breakable, 
  enhanced,
  sharp corners
]
\textit{Paper text: \{text\}}

\textbf{For these claims:}
\texttt{\{claims\_text\}}

\textbf{Please identify relevant evidence that:}
\begin{enumerate}
    \item Directly supports or contradicts the claim's specific assertion.
    \item Is presented with experimental results, data, or concrete examples.
    \item Can be traced to specific methods, results, or discussion sections.
    \item Is not from the abstract or introduction.
\end{enumerate}

\textbf{Return ONLY the following JSON:}
\begin{lstlisting}
{
    "evidence_sets": [
        {
            "claim_id": number,
            "evidence": [
                {
                    "evidence_id": number,
                    "evidence_text": "specific evidence",
                    "strength": "strong/moderate/weak",
                    "limitations": "key limitations",
                    "location": "section/paragraph",
                    "exact_quote": "verbatim text"
                }
            ]
        }
    ]
}
\end{lstlisting}
\end{tcolorbox}

% \newpage

\begin{tcolorbox}[
  title=Conclusion Evaluation Prompt, 
  colback=white, 
  colframe=green!75!black, 
  fonttitle=\bfseries, 
  breakable, 
  enhanced,
  sharp corners
]
\textbf{Analyze these claims and their evidence:}
\texttt{\{analysis\_text\}}

\textbf{For each claim-evidence pair, evaluate:}
\begin{enumerate}
    \item Whether the evidence justifies the claim.
    \item The overall strength of support.
    \item Any important limitations.
\end{enumerate}

\textbf{Return ONLY the following JSON:}
\begin{lstlisting}
{
    "conclusions": [
        {
            "claim_id": number,
            "conclusion_justified": true/false,
            "robustness": "high/medium/low",
            "key_limitations": "specific limitations",
            "confidence_level": "high/medium/low"
        }
    ]
}
\end{lstlisting}
\end{tcolorbox}

\vspace{6mm}

\subsection{One-by-One Prompt}
\label{1_by_1_prompt}

\begin{tcolorbox}[
  title=Claims Extraction Prompt, 
  colback=white, 
  colframe=blue!75!black, 
  fonttitle=\bfseries, 
  breakable, 
  enhanced, % ensures enhanced features are active
  sharp corners
]
Analyze this research paper and extract ALL possible claims made by the authors.
\textit{Paper text: \{text\}}

Your task is to identify all statements in the text that meet the following criteria for a claim:
\begin{enumerate}
    \item Makes a specific, testable assertion about results, methods, or contributions.
    \item Represents a novel finding, improvement, or advancement.
    \item Presents a clear position or conclusion.
\end{enumerate}

Make sure to:
\begin{itemize}
    \item Include both major and minor claims.
    \item Don't miss any claims.
    \item Present each claim as a separate item.
\end{itemize}

Return ONLY the following JSON structure:
\begin{lstlisting}
{
    "claims": [
        {
            "claim_id": 1,
            "claim_text": "statement of the claim",
            "location": "section/paragraph where this claim appears",
            "claim_type": "Nature of the claim",
            "exact_quote": "complete verbatim text containing the claim"
        }
    ]
}
\end{lstlisting}
\end{tcolorbox}

% \newpage

\begin{tcolorbox}[
  title=Evidence Analysis Prompt, 
  colback=white, 
  colframe=red!75!black, 
  fonttitle=\bfseries, 
  breakable, 
  enhanced,
  sharp corners
]
\textit{Paper text: \{text\}}

For the following claim from the paper:
\texttt{"\{claim['claim\_text']\}"}

Please identify relevant evidence that:
\begin{enumerate}
    \item Directly supports or contradicts the claim's specific assertion.
    \item Is presented with experimental results, data, or methodology.
    \item Can be traced to specific methods, results, or discussion sections.
    \item Is not from the abstract or introduction.
\end{enumerate}

If NO evidence is found for the given Claim, return:
\begin{lstlisting}
{
    "claim_id": {claim['claim_id']},
    "evidence": [],
    "no_evidence_reason": "Explain why no evidence was found (e.g., 'Claim is unsupported', 'Claim is theoretical without empirical evidence', etc.)"
}
\end{lstlisting}
ELSE:
Return ONLY the following JSON structure:
\begin{lstlisting}
{
    "claim_id": {claim['claim_id']},
    "evidence": [
        {
            "evidence_id": 1,
            "evidence_text": "specific experimental result/data point",
            "evidence_type": "primary/secondary",
            "strength": "strong/moderate/weak",
            "limitations": "stated limitations or assumptions",
            "location": "specific section & paragraph",
            "exact_quote": "verbatim text from paper"
        }
    ]
}
\end{lstlisting}
\end{tcolorbox}

% \newpage

\begin{tcolorbox}[
  title=Conclusion Analysis Prompt, 
  colback=white, 
  colframe=green!75!black, 
  fonttitle=\bfseries, 
  breakable, 
  enhanced,
  sharp corners
]
\textit{Paper text: \{text\}}

Analyze the following claim and its supporting evidence:
\texttt{\{single\_claim\_analysis\}}

Provide a comprehensive conclusion analysis following these guidelines:
\begin{enumerate}
    \item Evidence Assessment:
    \begin{itemize}
        \item Evaluate the strength and quality of ALL evidence presented.
        \item Consider both supporting and contradicting evidence.
        \item Assess the methodology and reliability of evidence.
    \end{itemize}

    \item Conclusion Analysis:
    \begin{itemize}
        \item Determine what the authors concluded about this specific claim.
        \item Evaluate if the conclusion is justified by the evidence.
        \item Consider the relationship between evidence quality and conclusion strength.
    \end{itemize}

    \item Robustness Evaluation:
    \begin{itemize}
        \item Assess how well the evidence supports the conclusion.
        \item Consider methodological strengths and weaknesses.
        \item Evaluate the consistency of evidence.
    \end{itemize}

    \item Limitations Analysis:
    \begin{itemize}
        \item Identify specific limitations in both evidence and conclusion.
        \item Consider gaps in methodology or data.
        \item Note any potential biases or confounding factors.
    \end{itemize}
\end{enumerate}

Return ONLY the following JSON structure:
\begin{lstlisting}
{
    "conclusions": [
        {
            "claim_id": {claim_id},
            "author_conclusion": "detailed description of authors' conclusion based on evidence",
            "conclusion_justified": true/false,
            "justification_explanation": "detailed explanation of why conclusion is/isn't justified",
            "robustness_analysis": "comprehensive analysis of evidence strength and reliability",
            "limitations": "specific limitations and caveats",
            "location": "section/paragraph where conclusion appears",
            "evidence_alignment": "analysis of how well evidence aligns with conclusion",
            "confidence_level": "high/medium/low based on evidence quality"
        }
    ]
}
\end{lstlisting}
\end{tcolorbox}
\newpage

\section{Additional Details on Annotation}
\subsection{Annotator Guidelines}
\label{Annotator_guidelines}
\begin{itemize}
\item Select one recent research paper in the field of artificial intelligence or machine learning.
\item Prioritize papers published in 2024 to ensure relevance to current developments.
\item When possible, select a paper with fewer than 20 pages to facilitate thorough annotation.
\item Avoid papers with heavily mathematical content to ensure accessibility.
\item Complete all annotation tasks independently, without employing large language models for assistance at any stage of the 
process.
\end{itemize}

\textbf{Task Description}

Your task is to identify all statements in the text that qualify as claims under the following criteria:

\begin{enumerate}
    \item \textbf{Specificity}: The statement makes a specific, testable assertion about results, methods, or contributions.
    \item \textbf{Novelty}: The statement represents a novel finding, improvement, or advancement.
    \item \textbf{Clarity}: The statement presents a clear position or conclusion.
\end{enumerate}

\textbf{Requirements}

\begin{itemize}
    \item Include both major and minor claims.
    \item Ensure no claim is overlooked.
    \item Present each claim as a separate item.
\end{itemize}

\textbf{Evidence Identification}

For each identified claim, find and document relevant evidence that:

\begin{enumerate}
    \item \textbf{Relevance}: Directly supports or contradicts the claim's specific assertion.
    \item \textbf{Concrete Support}: Is presented with experimental results, data, or concrete examples.
    \item \textbf{Traceability}: Can be traced to specific methods, results, or discussion sections in the text.
    \item \textbf{Exclusions}: Evidence must not be derived from the abstract or introduction sections of the text.
\end{enumerate}

\textbf{Conclusion Analysis}

\begin{itemize}
    \item \textbf{Justification}: Evaluate whether the conclusions drawn in the text are justified by the evidence provided.
\end{itemize}

\textbf{Annotation Format}

Each annotation should be formatted as follows:

\begin{verbatim}
{
    "Claim_id": "<unique_identifier>",
    "Claim_text": "<text_of_the_claim>",
    "Evidence_text": "<text_supporting_or_contradicting_the_claim>",
    "Justification_Conclusion": "<evaluator's_comment_on_evidence_justification>"
}
\end{verbatim}

% \newpage

\subsection{Inter-Annotator Agreement Methodology}
\label{inter_annotator_agreement}
To evaluate the reliability of the CLAIM-BENCH annotations, we calculated Inter-Annotator Agreement on a subset of 30 papers, each annotated by two different annotators on the Claims and the Evidence. For each of the claims and the evidences, we take one set (``set A'') as the ground truth and compute the F1-score of the other set (``set B''). Considering the symmetry, we also computed the F1-score swapping sets A and B, and reported the averaged F1-score. We chose F1 because our annotation task (identifying and linking spans) closely parallels standard information extraction tasks, where F1 is a standard evaluation measure balancing precision and recall; this reflects the need for agreement on both the correctness and comprehensiveness of annotations.

% \paragraph{Inter-Annotator Agreement (IAA).}
Apart from this we also used an LLM assistant (Gemini 2.5) to automate Cohen’s~$\kappa$ on a 30-paper subset.
For every paper the LLM (i) extracted the two raw annotation files, (ii) built binary vectors of length $N$ (one entry per sentence; 1 = tagged, 0 = untagged) for \textit{claims} and for \textit{evidence}, (iii) populated the $2{\times}2$ contingency table $(a,b,c,d)$, and (iv) computed
$\kappa=\tfrac{p_o-p_e}{1-p_e}$.
the procedure was spot-checked on 10 papers and the LLM’s arithmetic matched manual counts exactly.

The aggregated results are \textbf{$\kappa=0.66$ for claims} (substantial agreement) and \textbf{$\kappa=0.30$ for evidence} (fair agreement).
The lower evidence score is expected: evidence sentences are sparse ($<0.3\%$ of text) and dispersed, so chance agreement is already high, and even minor boundary or selection differences depress $\kappa$.
Moreover, a single claim can legitimately map to several evidence sentences; annotators often choose different yet valid spans, further reducing overlap.
Despite this, both scores confirm that CLAIM-BENCH offers a dependable—though challenging—ground-truth resource for benchmarking claim–evidence reasoning.

\begin{tcolorbox}[
  title=\textbf{Cohen’s $\kappa$ Agreement Prompt},
  colback=white,
  colframe=green!75!black,
  fonttitle=\bfseries,
  breakable,
  enhanced,
  sharp corners
]
\textit{Paper filename: \{pdf\_name\}}  
Total sentences in paper: \texttt{\{total\_sentences\}}

\bigskip
You are given two raw annotation lists for \emph{claim identification}—one from
Annotator 1 and one from Annotator 2.  Follow the steps below \textbf{exactly}
to compute \textbf{Cohen’s $\kappa$}:

\begin{enumerate}
    \item \textbf{Vector Construction}  
          Build two binary vectors of length $N = \{total\_sentences\}$:  
          \begin{itemize}
              \item \texttt{1} if the sentence was marked as a claim by the annotator.  
              \item \texttt{0} if the sentence was \emph{not} marked as a claim.
          \end{itemize}

    \item \textbf{Contingency Table}  
          Using the two vectors, populate the $2 \times 2$ table:
          \[
          \begin{array}{c|cc}
              & \text{Ann 2} = 1 & \text{Ann 2} = 0\\ \hline
              \text{Ann 1} = 1 & a & b\\
              \text{Ann 1} = 0 & c & d
          \end{array}
          \]

    \item \textbf{Compute $\kappa$}  
          \begin{align*}
              P_o &= \frac{a + d}{N} \\[2pt]
              P_e &= \Bigl(\frac{a+b}{N}\Bigr)\Bigl(\frac{a+c}{N}\Bigr) \;+\;
                    \Bigl(\frac{c+d}{N}\Bigr)\Bigl(\frac{b+d}{N}\Bigr) \\[2pt]
              \kappa &= \frac{P_o - P_e}{1 - P_e}
          \end{align*}

    \item \textbf{Return \emph{only} the JSON below}:  
\begin{lstlisting}
{
    "kappa_claims": 0.00
}
\end{lstlisting}
\end{enumerate}

\bigskip
\textbf{Raw Annotations – Annotator 1:} \{\texttt{raw\_annotations1}\}

\medskip
\textbf{Raw Annotations – Annotator 2:} \{\texttt{raw\_annotations2}\}
\end{tcolorbox}

% \newpage

\begin{tcolorbox}[
  title={Example Output: Cohen’s $\kappa$ Calculation},
  colback=white,
  colframe=green!75!black,
  fonttitle=\bfseries,
  breakable,
  enhanced,
  sharp corners
]
We compute Cohen’s $\kappa$ for \textbf{claim identification} on a paper with
$N=667$ sentences.

\paragraph{Annotation statistics}
\begin{itemize}
  \item Annotator~1 marked \textbf{5} sentences as claims.
  \item Annotator~2 marked \textbf{6} sentences as claims.
  \item Overlap (both $\mathit{claim}=1$): \textbf{4} sentences.
\end{itemize}

\paragraph{Contingency table}
\[
\begin{array}{c|cc|c}
 & \text{Ann\,2}=1 & \text{Ann\,2}=0 & \text{Row Tot.} \\ \hline
\text{Ann\,1}=1 & 4 & 1 & 5 \\
\text{Ann\,1}=0 & 2 & 660 & 662 \\ \hline
\text{Col. Tot.} & 6 & 661 & 667
\end{array}
\]

\paragraph{Calculations}
\begin{align*}
P_o &= \frac{a+d}{N} \;=\; \frac{4+660}{667} \;\approx\; 0.9955,\\[4pt]
P_e &= 
\Bigl(\frac{a+b}{N}\Bigr)\Bigl(\frac{a+c}{N}\Bigr) \;+\;
\Bigl(\frac{c+d}{N}\Bigr)\Bigl(\frac{b+d}{N}\Bigr) \\[2pt]
&= \Bigl(\tfrac{5}{667}\Bigr)\Bigl(\tfrac{6}{667}\Bigr) +
   \Bigl(\tfrac{662}{667}\Bigr)\Bigl(\tfrac{661}{667}\Bigr)
 \;\approx\; 0.98375,\\[8pt]
\kappa &= \frac{P_o - P_e}{1 - P_e}
       \;=\;
       \frac{0.99550 - 0.98375}{1 - 0.98375}
       \;\approx\; 0.7231.
\end{align*}
\textbf{Result JSON}
% \paragraph{}

\begin{lstlisting}
{
    "kappa_claim": 0.7231
}
\end{lstlisting}

\end{tcolorbox}

% \newpage
\subsection{Annotation Tool}
\label{annotation_tool}
\begin{figure*}[!h]
    \centering
    \includegraphics[width=\textwidth]{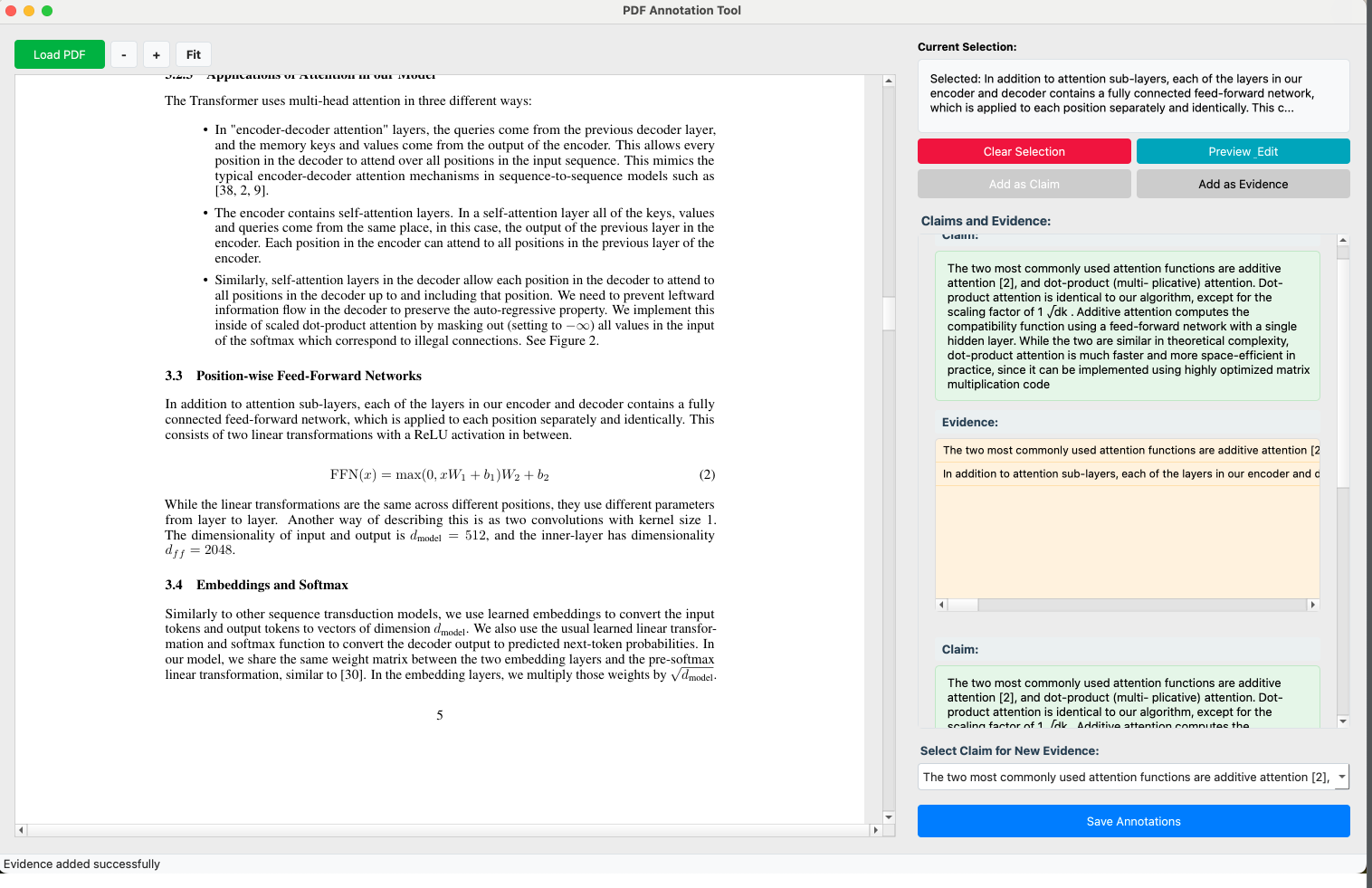}
    \caption{The custom annotation tool interface used for CLAIM-BENCH dataset creation, enabling direct PDF text selection and structured labeling (e.g., 'Add as Claim' button) of claim-evidence pairs.}

    \label{fig:annotation_tool}
\end{figure*}

\newpage

\section{Impact of Documents' Token Length}
\label{token_length}
\begin{figure*}[!t]
    \centering
    % First row
    \begin{subfigure}[t]{0.49\textwidth}
        \centering
        \includegraphics[width=\linewidth]{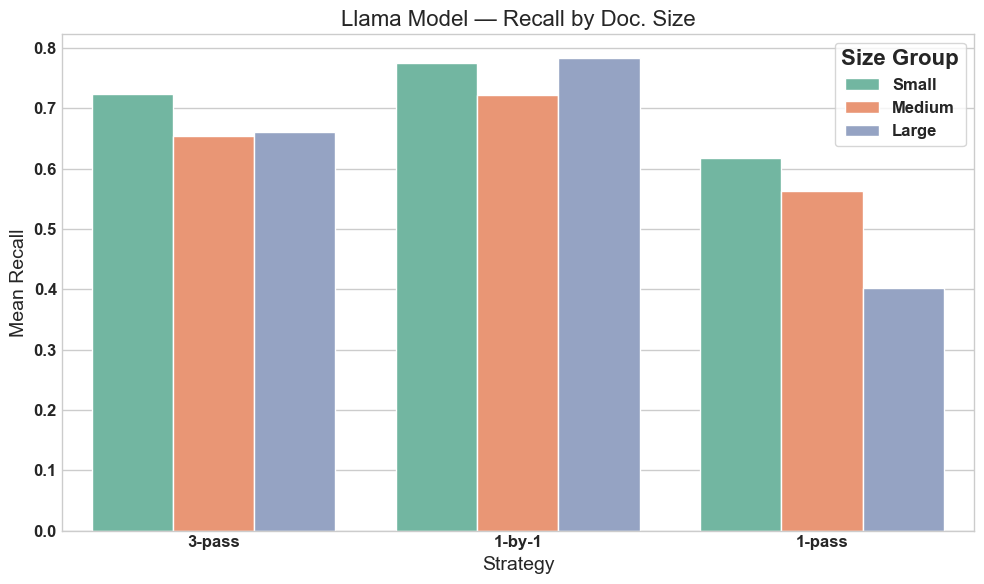}
        \caption{LLAMA Recall}
        \label{fig:llama_recall}
    \end{subfigure}
    \hfill
    \begin{subfigure}[t]{0.49\textwidth}
        \centering
        \includegraphics[width=\linewidth]{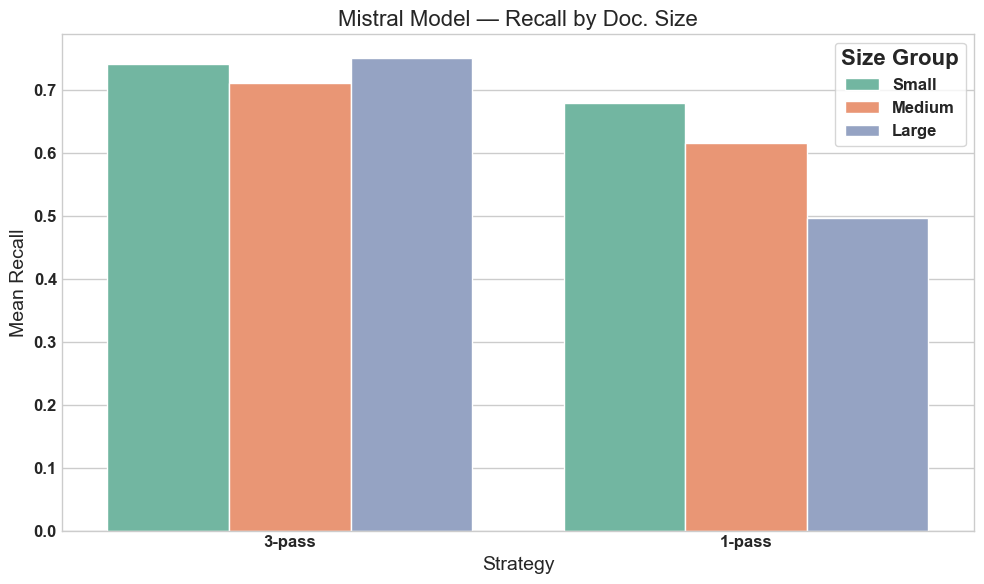}
        \caption{Ministral Recall}
        \label{fig:Mistal_recall}
    \end{subfigure}

    \vspace{0.5cm} % vertical spacing between rows

    % Second row
    \begin{subfigure}[t]{0.48\textwidth}
        \centering
        \includegraphics[width=\linewidth]{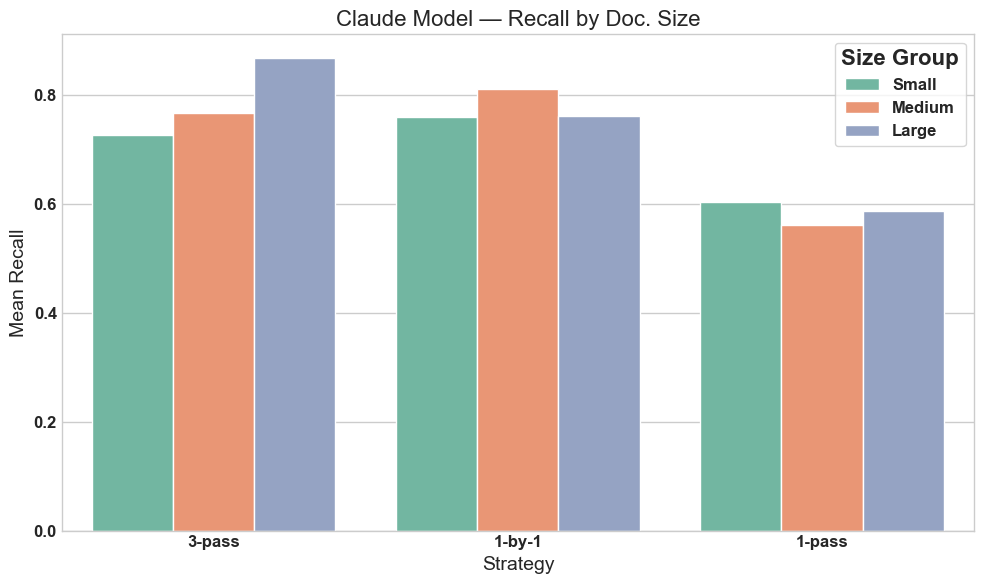}
        \caption{Claude Recall}
        \label{fig:image3}
    \end{subfigure}
    \hfill
    \begin{subfigure}[t]{0.48\textwidth}
        \centering
        \includegraphics[width=\linewidth]{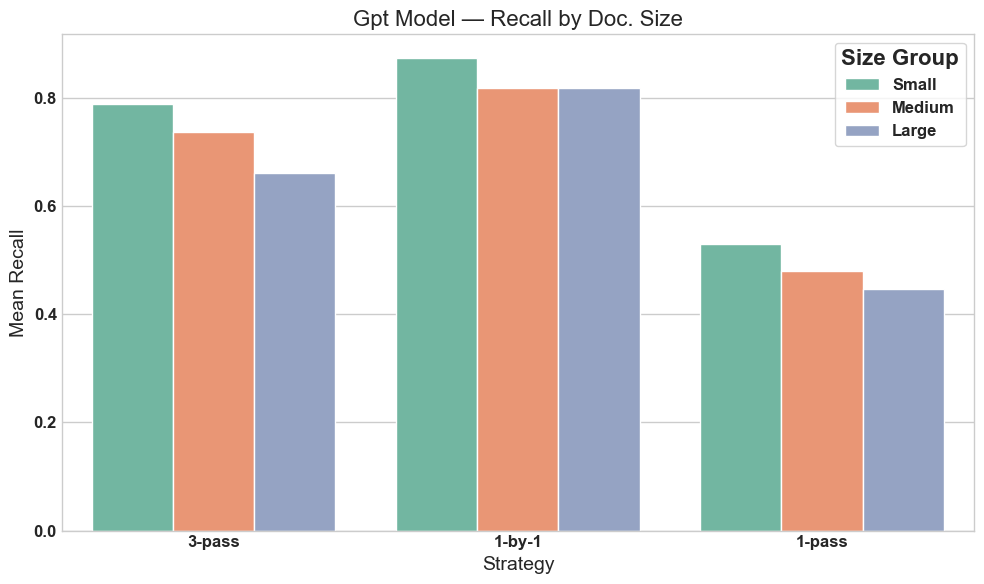}
        \caption{GPT-4 Recall}
        \label{fig:image4}
    \end{subfigure}

    \caption{Mean recall by document size groups (small, medium, large) for different models and prompting strategies, illustrating performance trends across increasing token counts.}
    \label{fig:comapre_recall}
\end{figure*}

% Appendix Figure A.1 plots mean recall against average token-count bins for each prompting strategy and lets us

Figure \ref{fig:comapre_recall} plots mean recall for three prompting strategies---Three-Pass, One-by-One, and Single-Pass---across three document-length buckets ($<$ 15 k, 15--20 k, $\geq$ 20 k tokens). A closer reading of the bars yields three key observations:

\begin{enumerate}
    \item \textbf{Performance drops are tied to the strategy more than the model size.}
    \begin{itemize}
        \item For every model, the Single-Pass run shows the steepest decline as documents grow.
        \item Example: LLaMA's recall plunges from about 0.60 in small papers to roughly 0.40 in $\geq$20 k-token papers under Single-Pass.
    \end{itemize}

    \item \textbf{Once an iterative strategy is used, the size-related gap all but disappears.}
    \begin{itemize}
        \item Iterative prompting (Three-Pass or One-by-One) largely neutralises length effects---even for the smaller models.
        \item LLaMA 70B: In One-by-One mode the large-document group matches or exceeds the small-document group ($\approx$ 0.78 vs $\approx$ 0.76).
        \item Ministral 8B: Three-Pass recall stays virtually flat ($\sim$ 0.72--0.75) across all three size buckets; the length penalty only appears in Single-Pass.
    \end{itemize}

    \item \textbf{Larger models still benefit, but their advantage is greatest with fine-grained prompts.}
    \begin{itemize}
        \item Claude 3.5 Sonnet: Recall rises with document size under Three-Pass ($\approx$ 0.72 $\rightarrow$ 0.85), and remains $\geq$ 0.75 in One-by-One.
        \item GPT-4-Turbo: One-by-One keeps recall at or above 0.80 for medium- and large-size papers; the drop to $\sim$ 0.66 for large papers occurs only in Three-Pass, not in Single-Pass.
    \end{itemize}
\end{enumerate}

The figure shows that prompt granularity is the dominant lever for long-context recall. Single-pass prompting amplifies context-window limits---especially in smaller models---but iterative, claim-level prompting (Three-Pass and One-by-One) recovers performance, sometimes even improving it as the text grows. Larger models are naturally more stable, yet they, too, realise their full potential only when given finer-grained, multi-step instructions.

\subsection{Sentence Distance Detailed Analysis}
\begin{figure*}[!h]
    \centering
    \includegraphics[width=\textwidth]{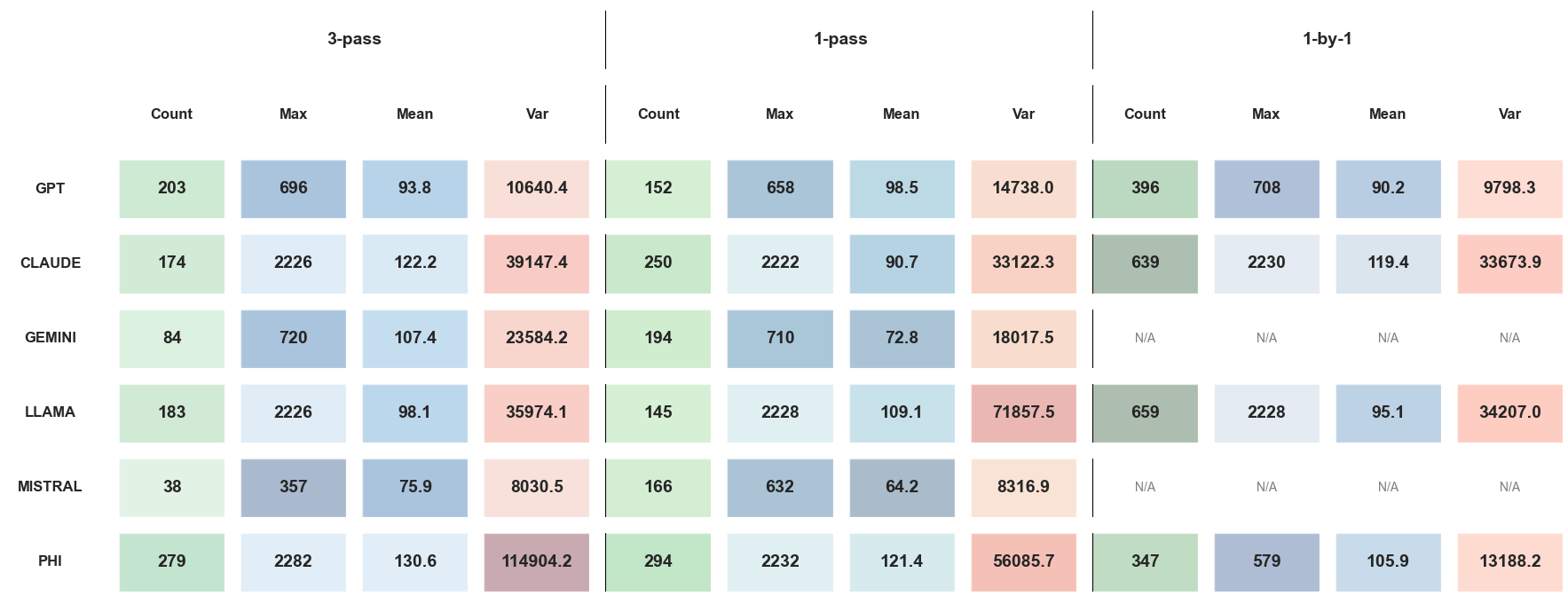}
    \caption{Aggregated statistics of the sentence\_gap metric Count, Max, Mean, and Variance (Var)—for each model under the three prompting strategies (Three-Pass, One-pass, and One-by-One). Larger counts and wider gaps (e.g., Claude and LLaMA exceeding 2,200-sentence links in One-by-One) reflect broader retrieval, whereas smaller models such as Ministral keep distances short and variance low. “N/A” indicates the model-strategy combination was not executed.}

    \label{fig:Sentence_distance_analysis}
\end{figure*}

\end{document}